\documentclass[11pt,a4paper]{article}

\usepackage[utf8]{inputenc}
\usepackage[margin=1in]{geometry}
\usepackage{amsmath, amssymb, amsfonts}
\usepackage{graphicx}
\usepackage{siunitx}
\usepackage{booktabs}
\usepackage{multirow}
\usepackage{authblk}
\usepackage[hidelinks]{hyperref}
\usepackage{caption}
\usepackage{subcaption}
\usepackage[numbers,sort&compress]{natbib}
\usepackage{wrapfig}
\usepackage[normalem]{ulem}

\newcommand{\DT}{\mathrm{DT}}            
\newcommand{\DTp}{\widehat{\mathrm{DT}}} 

\title{\textbf{
Fiberpose: state-of-the-art fiber morphometry and instance segmentation with a distance-transform neural network
}}

\author{Viktor Balashov, Daniil S. Antonenko}

\date{}

\begin{document}
\maketitle

\noindent\textbf{Web tool:} \url{https://fiberpose.org}

\noindent\textbf{Code availability:} \url{https://github.com/vikba/fiberpose}

\noindent\textbf{Correspondence:} \href{mailto:fiberpose@gmail.com}{fiberpose@gmail.com}

\begin{abstract}
Analysis of fibrous structures in SEM images is a routine task in biomedical engineering and materials science. However, most available methods for fiber quantification still rely on classical image binarization followed by simple post-processing. These approaches often produce unreliable results in images with weak boundaries, heterogeneous contrast, and densely packed fibers. Here, we introduce \textit{Fiberpose}, a framework that reformulates fiber analysis as direct regression of a smooth distance-transform (DT) field, together with a novel set of geometry-based algorithms specifically adapted to this representation. The predicted field directly encodes fiber geometry, while the accompanying algorithms enable instance reconstruction and measurement of diameter and orientation within a unified workflow. The framework is implemented as a standalone Python desktop application and a web tool (\url{https://fiberpose.org}). On an independent dataset of fibrous SEM images, \textit{Fiberpose} achieved state-of-the-art instance-segmentation accuracy in comparison with \textit{SAM2} and \textit{Cellpose-SAM}, and more accurate diameter measurements than \textit{DiameterJ}, \textit{Fibar} and \textit{\textit{GIFT}}. The method remains effective in dense and weakly contrasted regions and shows qualitative generalization across microscopy modalities not represented in the training data. \textit{Fiberpose} therefore provides an interpretable geometric representation and an automated pipeline for reproducible quantitative characterization of fibrous materials. 
\end{abstract}

\section*{Introduction}
Microscopic analysis of fiber morphology is a fundamental and routine task in both the life and materials sciences. In biology, fibrous structures are found at every scale, from cytoskeletal filaments, through the collagen and elastin networks of the extracellular matrix, to the cellulose microfibrils of plant cell walls, and their organization shapes tissue mechanics and cell behavior. In engineering, fibers are equally widespread: textiles, paper, fiber-reinforced composites, filtration media, and electrospun scaffolds for tissue engineering all derive their strength, stiffness, permeability, and biological response from the arrangement of their fibers. Despite these different contexts, the same quantitative descriptors are needed: fiber alignment, diameter distribution, and porosity link structure to function, but they are useful only if they can be measured consistently across imaging modalities and spatial scales.

Over the past decade, more than twenty dedicated software tools have been developed to automate quantitative analysis of fibers across microscopy modalities  \cite{deVries2023, Easwaran2016, Tort2026, Murphy2020}. Despite differences in implementation, most methods rely on classical image-processing techniques, such as Otsu thresholding \cite{Otsu1979}, statistical region merging (SRM) \cite{Nock2004}, or edge detection, followed by post-processing steps such as skeletonization and contour analysis. For example, \textit{DiameterJ} \cite{Hotaling2015} and \textit{Gordium} \cite{DAmore2010} use binarization as an initial step, whereas \textit{GIFT} \cite{Huling2022} employs Sobel edge detection. While effective for clean, high-contrast images, these methods fail when confronted with low signal-to-noise ratios and variable illumination. A more fundamental limitation does not depend on image quality: binary semantic representations cannot distinguish individual fibers at contacts and intersections, introducing ambiguities that can propagate into artifacts during subsequent morphological analysis, particularly in densely packed networks with overlapping fibers. Thus, the main bottleneck of currently available fiber analysis methods lies not only in the quality of binarization but in the binary representation itself.

\begin{figure}[htb]
  \centering
  \includegraphics[width=\linewidth]{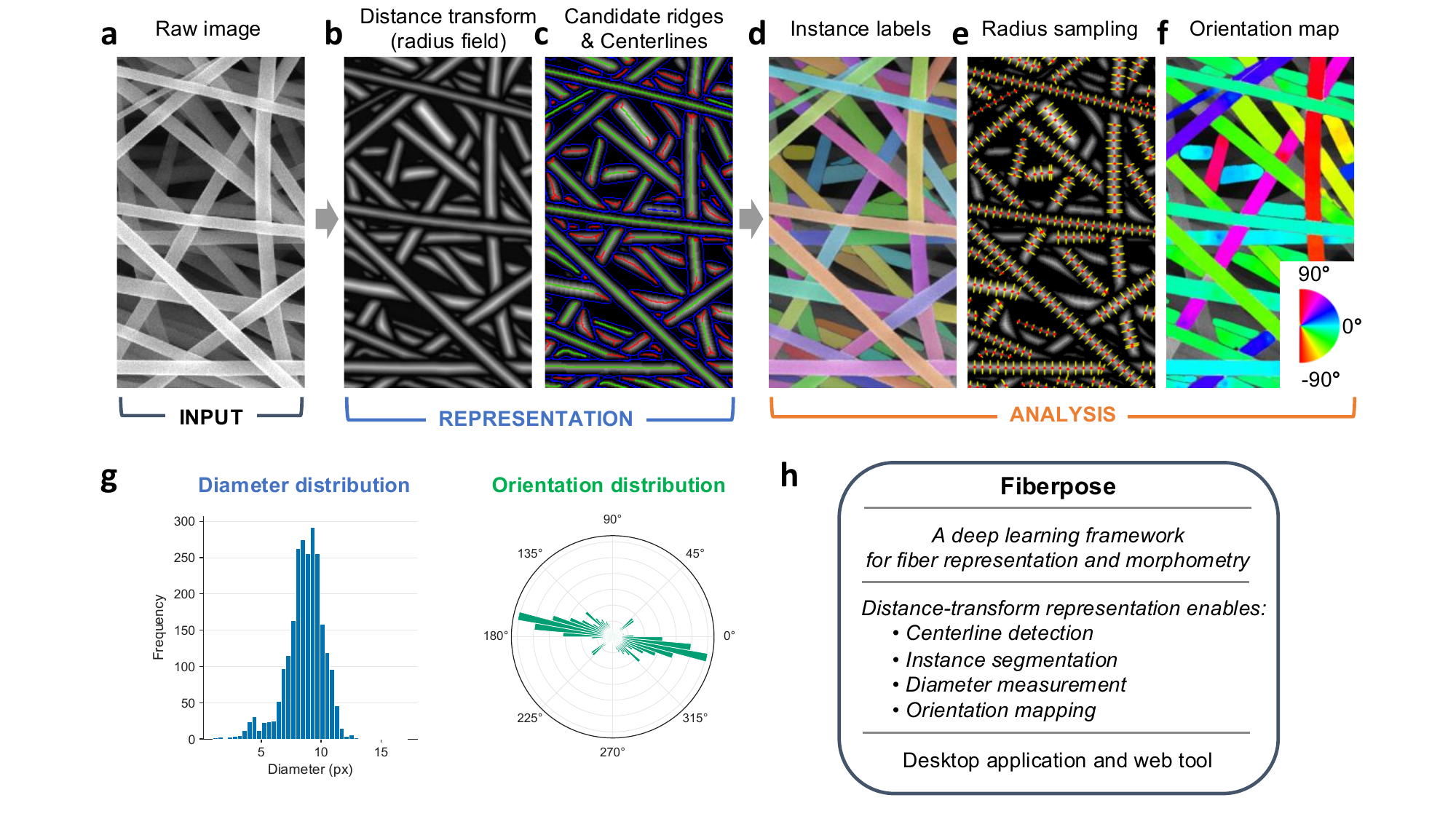}
  \caption{
  \textbf{Overview of the \textit{Fiberpose} pipeline for fiber network representation and morphometry.}
    (a) Raw microscopy image.
    (b, c) Geometric representation of the fiber network: (b) distance-transform (DT) field predicted by a neural network from the raw image, in which grayscale intensity represents the distance to the nearest fiber boundary; (c) candidate ridges (red) and measurement centerlines (green) traced on the DT field, with instance borders shown in blue.
    (d--f) Analysis steps: (d) fiber instance segmentation, with color denoting instance labels; (e) local radii sampled from the DT values along the measurement centerlines; (f) local orientations estimated from DT gradients near the measurement centerlines, color-coded according to the color wheel.
    (g) Diameter and orientation distributions within the sample.
    (h) Schematic summary of the \textit{Fiberpose} deep learning framework.
  \label{fig:pipeline}
  }
\end{figure}

Several deep-learning approaches have been proposed for fibrous structures, but most of them replace the binarization step without changing the representation. Semantic segmentation architectures, such as U-Net and DeepLabV3+ \cite{Nguyen2022, Kurkin2024}, substitute a learned classifier for thresholding, yet they produce the same kind of binary mask and therefore do not distinguish individual fibers at intersections or regions of overlap. Approaches that address fiber identity directly, such as Mask R-CNN and YOLOv8 instance segmentation \cite{Kurkin2024, Qamar2024} or U-Net-based segmentation followed by fiber tracing \cite{Peters2025}, have so far been demonstrated only on dispersed, individually resolvable filaments and are not suited to densely entangled networks containing numerous intersecting fibers. Interactive approaches, such as FiberDiameter \cite{Cossio_FiberDiameter}, enable measurements of selected fibers but are less suitable for automated whole-image analysis. Thus, learned segmentation, instance separation and fiber tracing have each been demonstrated, but only for dispersed fibers, and measurements are still derived from the resulting masks. What is still missing is a framework that provides both instance segmentation and quantitative morphometry for densely packed fibrous networks, across various materials, image qualities and fiber densities.

Recent advances in neural segmentation have shown that convolutional encoder--decoder architectures such as U-Net \cite{Ronneberger2015, Falk2019} can be adapted for instance segmentation in crowded scenes by predicting continuous geometric fields followed by post-processing, rather than relying solely on thresholded foreground probabilities. In cell image analysis, Cellpose \cite{Cellpose2021} and Omnipose \cite{Omnipose2022} predict a vector field for each image and recover instances by following it: pixels whose paths along the field converge to the same location are grouped into one object. In Cellpose, and in its extension with a vision-transformer backbone, Cellpose-SAM \cite{Pachitariu2025}, the field points toward cell centers, which becomes increasingly ambiguous for elongated structures and can lead to segmentation failures. In Omnipose, the field is the gradient of a distance field and points toward the medial axes of cells, enabling more robust segmentation of elongated and irregularly shaped instances. Yet Omnipose was developed and trained on images of bacteria and worms; the medial axes are never explicitly reconstructed or used as a geometric representation, and the framework provides no quantitative measurements of object geometry. For fibers and other objects with a high aspect ratio, however, the distance transform (DT) can serve as an explicit geometric representation in its own right: the same field that separates touching instances \cite{Naylor2019} also encodes the medial axis and local width of each object \cite{Wang2020}.

Here, we introduce \textit{Fiberpose}, a deep-learning-based framework that combines instance segmentation and quantitative morphometry of fibers through a common representation: the DT field and its ridges. It deliberately builds on two established ideas: the prediction of continuous geometric fields by a convolutional encoder--decoder network, as in Cellpose and Omnipose, and the distance transform that conventional fiber analysis tools, such as \textit{DiameterJ} \cite{Hotaling2015}, compute from a binarized image. \textit{Fiberpose} instead predicts the DT field directly from the original microscopy image, using a neural network trained on a curated dataset of diverse SEM images. This avoids error-prone binarization and enables a more reliable recovery of fiber geometry under challenging imaging conditions, including weak boundaries, heterogeneous contrast, and densely packed structures. Morphometry is performed directly on the predicted field and does not require successful instance segmentation. To exploit this representation, we developed a suite of geometric algorithms for centerline extraction, instance reconstruction, local diameter and orientation measurements, and assessment of measurement reliability. The contribution of this work therefore lies in the adaptation of individual components and their integration into a complete pipeline from the raw image to quantitative measurements, provided as freely available software through both a standalone desktop application and a web-based tool. \textit{Fiberpose} demonstrated higher measurement accuracy than the evaluated morphometry methods, as well as higher recall and precision than the evaluated instance segmentation methods.

\section*{Framework overview}

\subsection*{General architecture and terminology}

Fiberpose converts a raw microscopy image (Fig.~\ref{fig:pipeline}a, \textbf{INPUT}) into a geometric representation of the fiber network. First, a neural network predicts a distance-transform (DT) field, in which each foreground pixel encodes its estimated distance to the nearest fiber boundary. In this representation, fiber centerlines correspond to the ridges of the DT field, which reaches a local maximum along the direction perpendicular to the fiber. Fiberpose traces the candidate ridges and groups them into uniquely identified \textit{FiberFragments}. Geometrically derived rules are then applied to filter the candidate ridges, and the retained ones serve as measurement centerlines for quantitative morphometry (Fig.~\ref{fig:pipeline}b,c, \textbf{REPRESENTATION}).

This representation supports three \textbf{ANALYSIS} steps (Fig.~\ref{fig:pipeline}d--f). First, instance segmentation post-processes the DT field and candidate ridges to produce labeled fiber fragments.
In the second, independent step, Fiberpose obtains local fiber diameters. The methodology uses the fact that the value of the DT field at the ridge equals half the fiber thickness at that point. Therefore, by sampling twice the DT value along fiber centerlines at some spacing, Fiberpose estimates the distribution of local fiber diameters.
Third, local fiber orientation is estimated from DT gradients in the vicinity of centerline points. Since these gradients point across the fiber, the local fiber orientation is taken to be perpendicular to the gradient direction. The resulting orientations are interpolated within segmented fragments and color-coded to produce orientation maps.

Fragment-level measurements are aggregated into sample-level descriptors, including fiber-thickness distributions and orientation histograms (Fig.~\ref{fig:pipeline}g). This unified pipeline is implemented as an interactive analysis environment (Fig.~\ref{fig:pipeline}h) that links neural inference, geometric post-processing, visualization and quantitative export in a desktop application.

\subsection*{Predicting the distance transform (DT) field}

The network is trained to predict the DT, which, for a pixel $(x,y)$ belonging to fiber $k$ with region $\Omega_k$ and boundary $\partial\Omega_k$, is defined as
\begin{equation}
    DT(x,y) = \min_{(x',y')\in\partial\Omega_k} \sqrt{(x-x')^2+(y-y')^2}, \qquad (x,y)\in\Omega_k,
    \label{eq:DT-definition}
\end{equation}
and $DT=0$ for background pixels. This network, a compact U-Net that we call FPUNet, is trained on a curated dataset of scanning electron microscopy (SEM) images of fibers annotated by the authors.

A second, smaller FPUNet estimates foreground probability from the predicted DT field. This estimate can optionally be used to discard instances that do not correspond to fibers.

\begin{figure}[htb]
  \centering
  \includegraphics[width=0.9\linewidth]{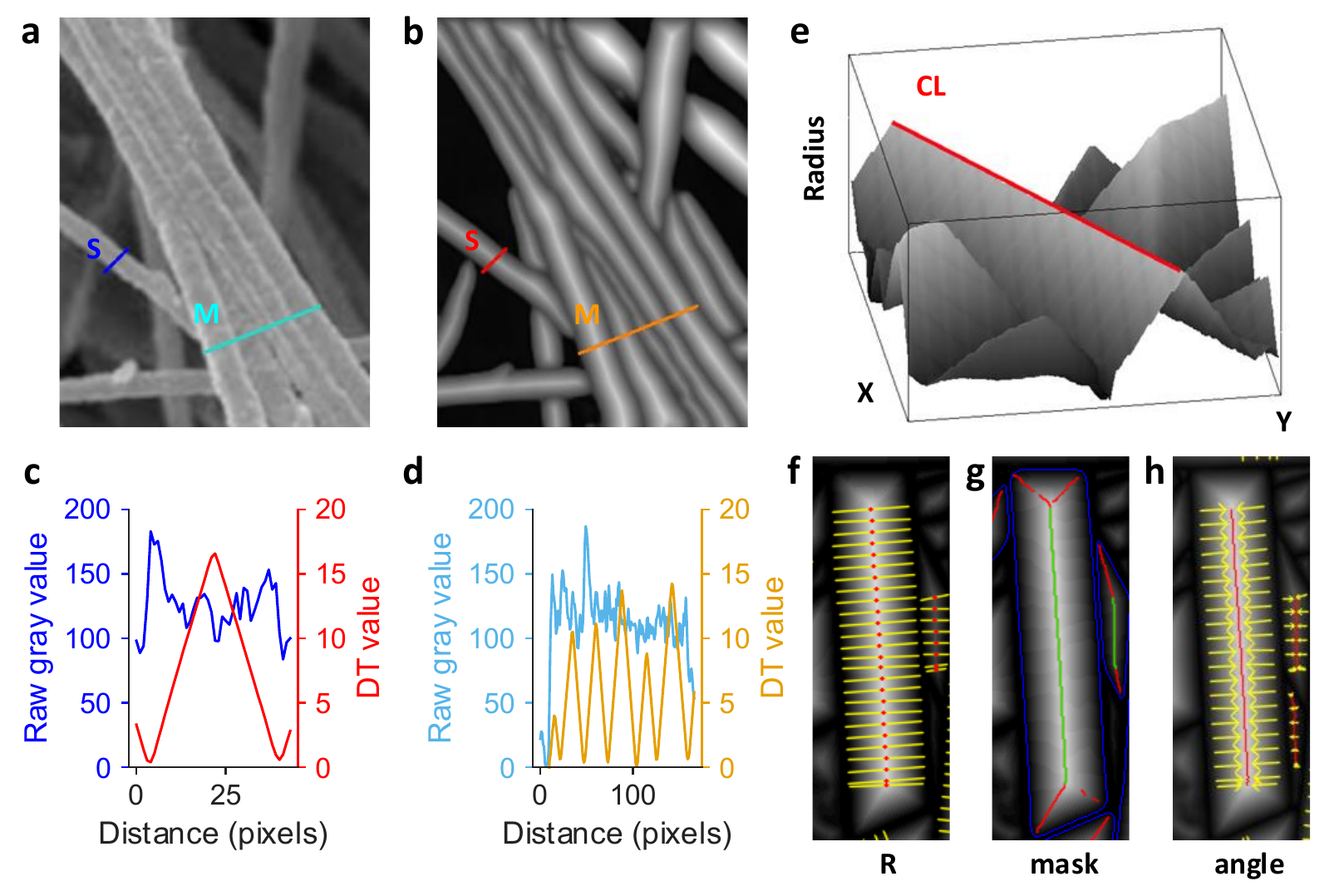}
  \caption{\textbf{Distance-transform representation enables robust fiber centerline detection and morphometry.} (a) Raw microscopy image. (b) predicted distance-transform (radius) field. Example cross sections across a single fiber (S, blue/red in panels a/b) and multiple adjacent fibers (M, cyan/orange in panels a/b). (c,d) Line profiles of raw grayscale intensity (blue, cyan) and distance-transform values (red/orange) along that cross sections. (e) Three-dimensional rendering of the distance-transform surface, with a detected centerline (CL, red) running along the ridge maximum. (f–h) Geometric parameters inferred along a centerline: local radius (f, R), fiber mask and centerline (g, mask) and local fiber orientation (h, angle).
  \label{fig:dt_details}
    }  
\end{figure}

\subsection*{DT representation of fibers}

A predicted DT field, unlike a binary mask or boundary map, encodes fiber geometry in a form that can be read out directly. For an ideal straight fiber of constant width, the DT cross-section is triangular, with a peak at the medial axis of the fiber. Its gradient has a unit magnitude everywhere within the fiber except on that axis, where the DT field exhibits a cusp. In practice, the predicted DT closely matches that theoretical shape: whereas the SEM intensity profile across a single fiber shows bright edge peaks separated by a noisy interior, the predicted profile is approximately triangular (Fig.~\ref{fig:dt_details}a--d). Across densely packed fibers, the predicted field resolves distinct ridges with triangular cross-sections for individual objects, even where boundaries are difficult to distinguish by eye in the raw image. This allows touching fibers to be separated at later stages of the pipeline.

\begin{figure}[htb]
  \centering
  \includegraphics[width=0.9\linewidth]{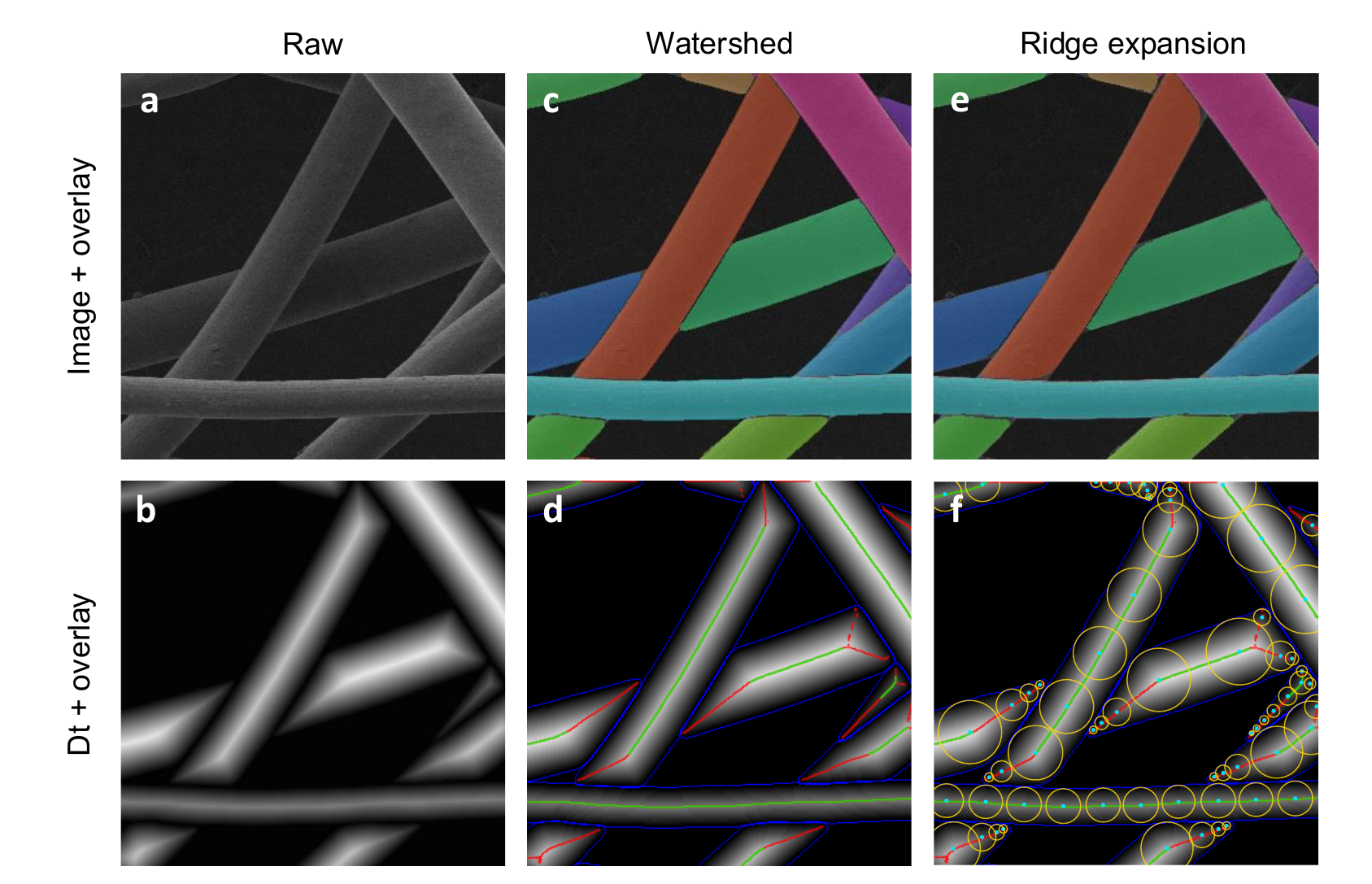}
  \caption{
  \textbf{Comparison of instance segmentation methods available in \textit{Fiberpose}: Watershed vs. Ridge expansion.}
(a) Original microscopy image \cite{Hotaling2015Dataset}. (b) Predicted distance-transform (DT) field.
(c) Marker-controlled watershed segmentation, showing segmented fiber instances. (d) Overlay of  candidate ridges (red), measurement centerlines (green), and segmentation boundaries (blue) on the DT field.
(e) Ridge expansion segmentation, showing segmented fiber instances. (f) Overlay of radial ridge-expansion regions (yellow circles), candidate ridges (red), measurement centerlines (green), and segmentation boundaries (blue) on the DT field. Compared to watershed segmentation, ridge expansion produces slightly rounded instance boundaries. 
}
  \label{fig:segmentation_mech}
\end{figure}

Viewed as a surface in three dimensions, the DT of each fiber forms an approximately triangular prism whose ridge follows the fiber axis (Fig.~\ref{fig:dt_details}e). Each analysis step reads a different property of this shape: ridge heights give local radii along measurement centerlines (Fig.~\ref{fig:dt_details}f); candidate ridges together with the surrounding field delimit individual instances (Fig.~\ref{fig:dt_details}g; Fig.~\ref{fig:segmentation_mech}); and gradients on the ridge flanks give the local normal and hence fiber orientation (Fig.~\ref{fig:dt_details}h). A single predicted field thus supports both instance reconstruction and direct quantitative measurement. Because the DT depends only on fiber geometry and not on imaging contrast, the downstream analysis is independent of image modality. The limiting factor is the quality of DT prediction, which depends on the modalities represented in the training data.

\subsection*{Tracing ridges and centerlines}

Because fiber centerlines are ridges of the DT field (Fig.~\ref{fig:dt_details}e), they can be traced with Steger's ridge detector~\cite{Steger1998}. The resulting candidate ridges include both fiber centerlines and V-shaped artifacts near fiber ends.

Candidate ridges are then grouped into \textit{FiberFragments}. Each ridge point is expanded into a disk with half its estimated local radius, and ridges whose disks overlap are assigned to the same fragment. This links connected structures, including fiber crossings and overlaps, while keeping distinct instances separate. Each fragment receives an identifier shared by its centerlines, measurements and segmented region, so that a \textit{FiberFragment} links the sampled diameters and orientations to their spatial locations and to the instance geometry for subsequent analysis.

Not every ridge point is suitable for measurement: fiber ends, crossings and uncertain regions distort the DT and are excluded from quantitative analysis. Measurement centerlines are therefore selected by geometric rules that test how closely the local DT matches that of an ideal fiber. By default, the local radius must vary little and change only gradually along the ridge. In the optional strict mode, the DT gradients on opposite sides of the centerline must also be antiparallel, and the DT must decrease away from the ridge with unit slope. Segments that pass these rules form the measurement centerlines; the remaining candidate ridges are still used for instance segmentation.

\subsection*{Fiber instance segmentation}

Fiber instances are built from the DT field and the candidate ridges using two alternative methods (Fig.~\ref{fig:segmentation_mech}). In marker-controlled watershed segmentation \cite{Beucher2018}, the DT is inverted so that ridges become basins, and each fragment's ridges serve as markers from which flooding proceeds until neighboring regions meet or the fiber edge is reached (Fig.~\ref{fig:segmentation_mech}c,d). Flooding is restricted to pixels with DT slope above desired minima. Ridge expansion segmentation instead takes the union of disks of local radius centered on each fragment's ridge points; it is simpler but produces slightly rounded instance boundaries (Fig.~\ref{fig:segmentation_mech}e,f). In both methods, instances are optionally cleaned up by removing very small, the lowest 1\% by centerline length, fragments and fragments with low predicted foreground probability.

\begin{figure}[h!]
    \centering
    \includegraphics[width=\textwidth]{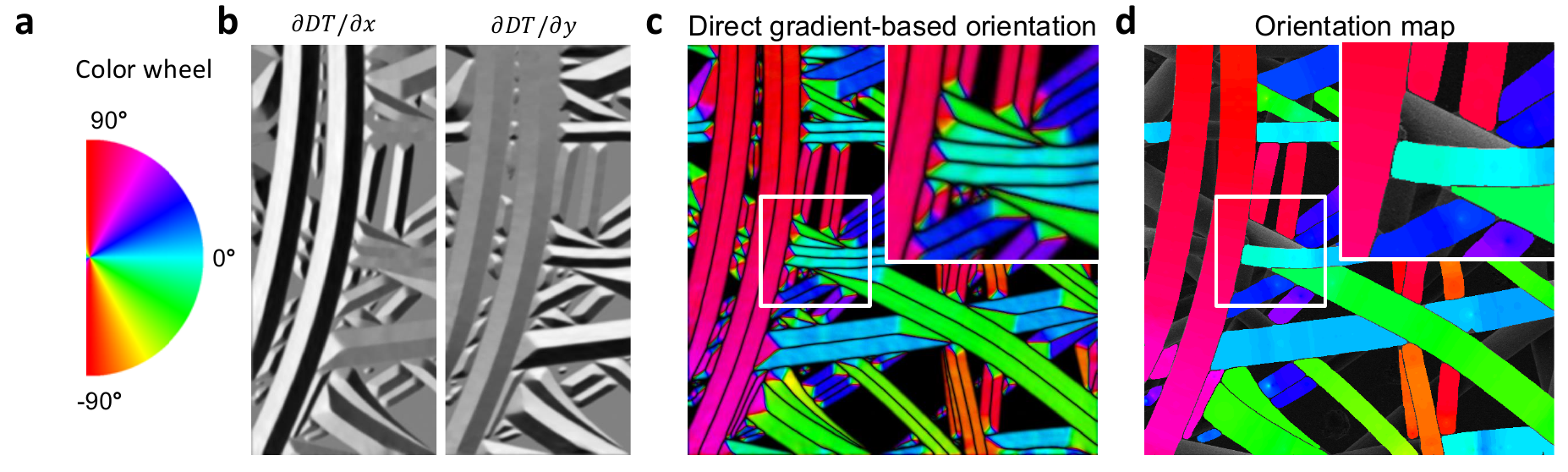}
    \caption{\textbf{Fiber orientation measurement from distance-transform gradients.}
    (a) Color wheel encoding local fiber orientations from $-90^\circ$ to $+90^\circ$.
    (b) Spatial derivatives of the DT along the $x$ and $y$ axis.
    (c) Direct gradient-based orientation map, with hue encoding the orientation perpendicular to the DT gradient and intensity encoding its magnitude. The inset magnifies the boxed region: dark lines run along DT ridges, where opposing gradients meet and the computed gradient magnitude decreases, and artifacts appear at fiber ends.
    (d) Resulting \textit{Fiberpose} map, obtained by sampling orientations along measurement centerlines and interpolating them within each segmented fragment using inverse-distance weighting. The inset shows the same region as in (c): coloring extends across fiber interiors without the ridge and fiber-end artifacts and is confined to the segmented regions.
     Source image from \cite{Hotaling2015Dataset}.}
    \label{fig:m_orientation}
\end{figure}

\subsection*{Diameter measurements}

Diameters are measured directly from the values of the predicted DT field along the measurement centerlines. FPUNet was trained with Eikonal regularization, which encourages a geometrically correct DT field everywhere except near ridges, where the DT is not differentiable. The predicted maxima are therefore left unconstrained and are usually slightly rounded (Fig.~\ref{fig:res_diameter_sampling}b), so the DT value on the centerline, the \emph{apex estimate}, underestimates the fiber radius. To reduce this effect, we developed a two-offset sampling procedure that takes DT values at a fixed offset on opposite sides of the centerline. For a centerline point $\mathbf{x}$ with unit normal $\mathbf{n}$, the \emph{corrected estimate} of the radius is

\begin{equation}
  \hat{r}(\mathbf{x}) = \tfrac{1}{2}\left[D(\mathbf{x} + \delta\mathbf{n}) + D(\mathbf{x} - \delta\mathbf{n})\right] + \delta,
  \label{eq:two_offset}
\end{equation}
where $D$ is the predicted DT field and $\delta$ is the sampling offset. The correction exploits the approximately linear transverse profile of the DT away from the ridge: for an ideal distance field, $D(\mathbf{x} \pm \delta\mathbf{n}) = r - \delta$, so the mean of the two flank samples plus $\delta$ recovers the radius even when the maximum itself is rounded. The local diameter is then $d(\mathbf{x}) = 2\hat{r}(\mathbf{x})$. The correction is evaluated against manual measurements in Results.

\begin{figure}[htb]
  \centering
  \includegraphics[width=0.9\linewidth]{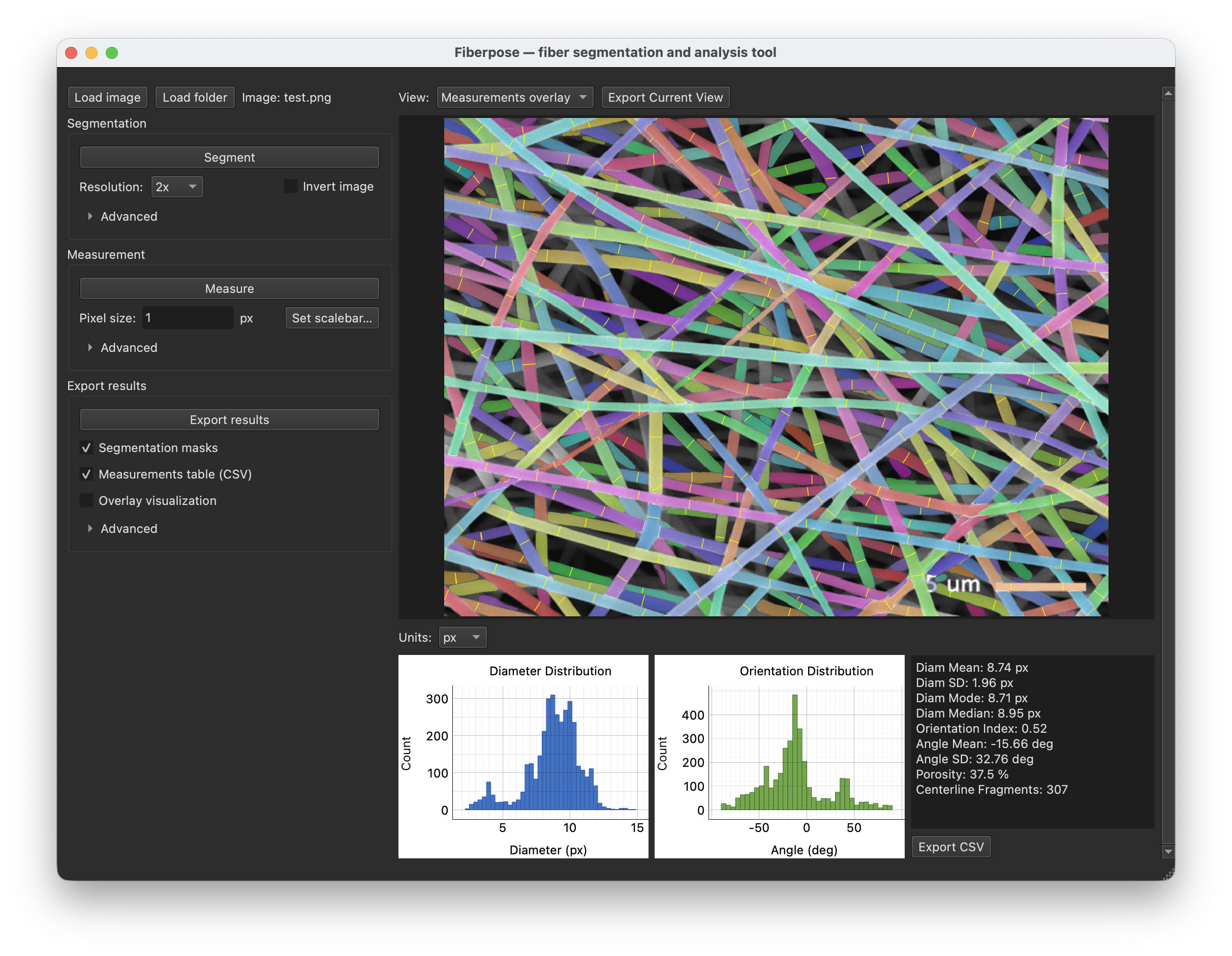}
  \caption{
  \textbf{\textit{Fiberpose} graphical user interface.}
  Screenshot of the interface, showing an overlay of instance segmentation and diameter measurements, together with sample-level distributions of fiber diameters and orientations.}
  \label{fig:GUI}
\end{figure}

\subsection*{Orientation analysis}

Fiber orientation is estimated from the DT gradient. The latter is perpendicular to the local fiber orientation, so it should be rotated by $\pm90^\circ$ to obtain the fiber direction during sampling. Angles follow the convention and color coding of \textit{OrientationJ}~\cite{Puspoki2016}: they range from $-90^\circ$ to $+90^\circ$, measured relative to the horizontal axis, and are encoded with a cyclic color wheel (Fig.~\ref{fig:m_orientation}a). Because the gradient direction is undefined on the ridge itself, Fiberpose samples gradients in the vicinity of each measurement centerline point rather than directly on it. 

The distribution of fiber orientations across the image is summarized as an angular histogram and an orientation index ranging from 0 for an isotropic to 1 for a fully aligned fiber network:

\begin{equation}
\mathrm{OI}
=
\sqrt{
\left\langle\cos(2\theta)\right\rangle^2+
\left\langle\sin(2\theta)\right\rangle^2
}.
\end{equation}

The orientation distribution can also be visualized directly by computing the gradient direction from the spatial derivatives $\partial D/\partial x$ and $\partial D/\partial y$ at every pixel (Fig.~\ref{fig:m_orientation}b,c). Such direct gradient maps, however, show artifacts near ridges, where opposing gradients meet, as well as near fiber ends and in short fragments with undefined centerlines (Fig.~\ref{fig:m_orientation}c, inset). Fiberpose therefore interpolates the centerline orientations throughout each segmented fragment using inverse-distance weighting, producing coloring that spans the entire instance area and is free of these artifacts (Fig.~\ref{fig:m_orientation}d, inset).

\subsection*{The software}

Fiberpose is implemented as a desktop application written in Python, with a graphical user interface (GUI) for interactive analysis and a command-line mode for batch processing of image folders and stacks (Fig.~\ref{fig:GUI}). A web version is available at \url{https://fiberpose.org}. The GUI displays every intermediate product of the pipeline, including the predicted DT field, candidate ridges, measurement centerlines, segmented instances, sampling sites and orientation maps. This allows users to inspect each step and judge how far the resulting measurements can be trusted. Results can be exported as images and as tables at both the per-image and per-sample level.

\section*{Results}

\begin{figure}[!htbp]
  \centering
  \includegraphics[width=0.9\textwidth]{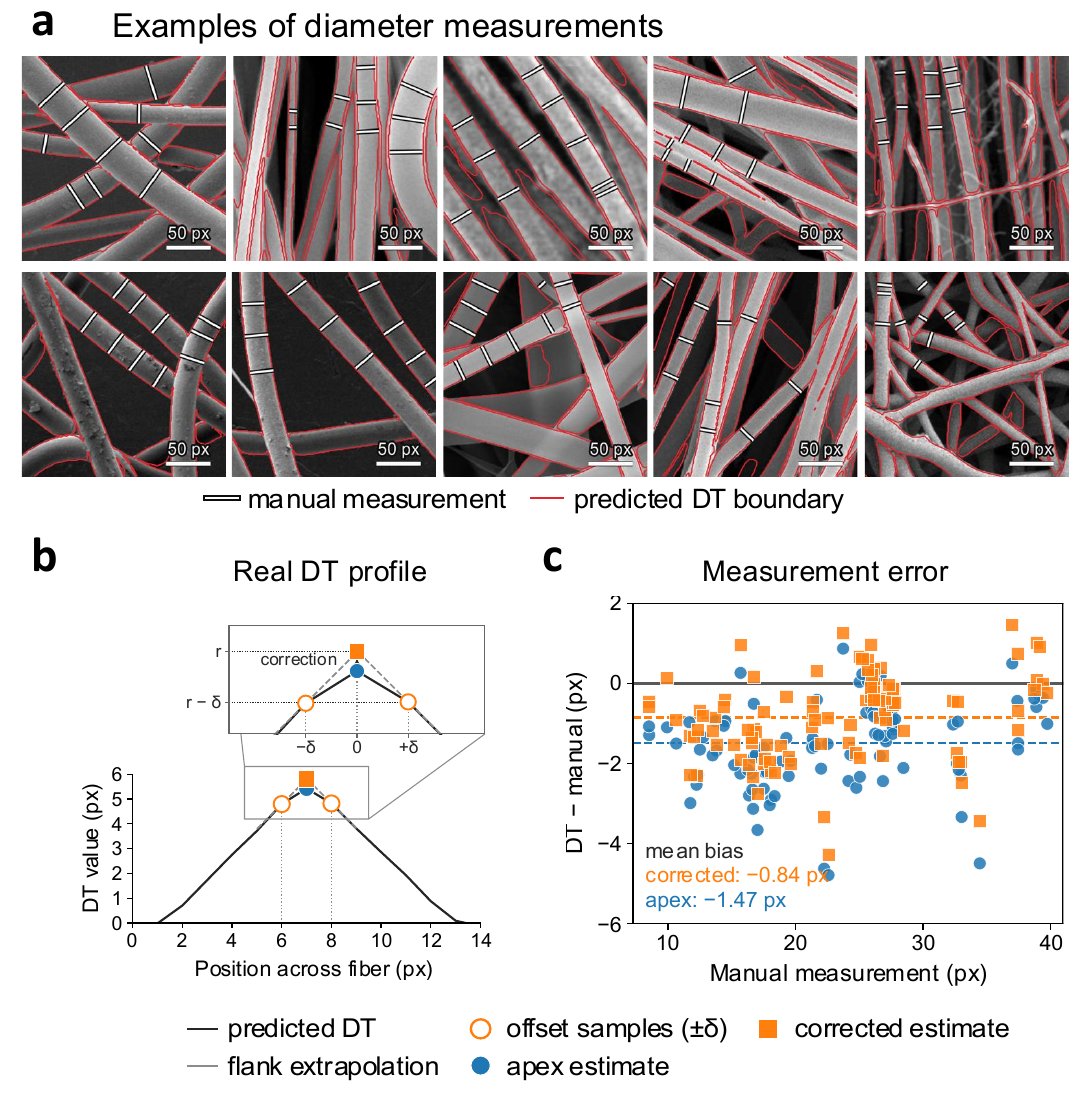}
  \caption{\textbf{The corrected estimate reduces the underestimation of fiber diameter by the apex estimate.}
  \textbf{(a)}~Representative regions of the ten validation SEM images (image sources: \cite{Hotaling2015Dataset, Huling2022, Sedlacek2023SEM, Roberto2022, Ciszewski2019, Bloise2018, Shaker2023}). White lines mark manual diameter measurements; red lines mark the boundary of the predicted distance-transform (DT) field ($\mathrm{DT} = 0.5$~px contour), from which automatic measurements are sampled. Ten lines were measured per image. Scale bars, 50~px.
  \textbf{(b)}~Representative DT profile across a single fiber. The close-up shows the apex in units of $\delta$. The apex estimate (blue circle) is the DT maximum. The offset samples (open circles) lie at $\pm\delta$ from the apex ($\delta = 1$~px). Dashed lines extend the flanks with unit slope through the offset samples and meet at the corrected estimate $r = \tfrac{1}{2}\left[D(+\delta) + D(-\delta)\right] + \delta$ (orange square). On an ideal DT, the offset samples lie one $\delta$ below $r$. The arrow marks the correction, from 5.42~px (apex) to 5.81~px (corrected).
  \textbf{(c)}~Difference between DT-derived and manual diameters as a function of the manual measurement, for the apex and corrected estimates. Dashed lines mark the mean bias. Both estimates were sampled at the nearest pixel: the apex estimate at the position of the maximal DT value along each manual line, and the offset samples at $\pm\delta$ from that position along the line. The corrected estimate reduced the mean absolute error from 1.53 to 1.07~px and the bias from $-1.47$ to $-0.84$~px, with lower absolute error in nine of the ten images (one-sided paired Wilcoxon signed-rank test, $n = 10$ images, $p = 0.002$).
}
  \label{fig:res_diameter_sampling}
\end{figure}

\subsection*{Correcting the apex estimate of fiber diameter}

We compared the apex and corrected estimates at identical, manually defined locations. Manual diameters (ten per image) were measured in ImageJ on ten SEM images not used for training.

The apex estimate systematically underestimated fiber diameter, with a mean bias of $-1.47$~px and a mean absolute error (MAE) of 1.53~px across the ten images (Fig.~\ref{fig:res_diameter_sampling}c). The corrected estimate reduced the bias to $-0.84$~px and the MAE to 1.07~px, a 29.9\% reduction, over the full range of measured diameters (9--40~px). The MAE was lower in nine of the ten images (one-sided paired Wilcoxon signed-rank test, $n = 10$ images, $p = 0.002$).

The residual bias of the corrected estimate remained different from zero (mean $-0.84$~px; 95\% CI over per-image mean biases $-1.32$ to $-0.37$~px, $n = 10$ images) and was negative in nine of the ten images. It should be read against the uncertainty of the manual reference: measurement lines drawn slightly obliquely to the fiber axis overestimate the diameter, so the manual values are themselves likely biased upward.

\begin{figure}[htb]
\centering
\includegraphics[width=\linewidth]{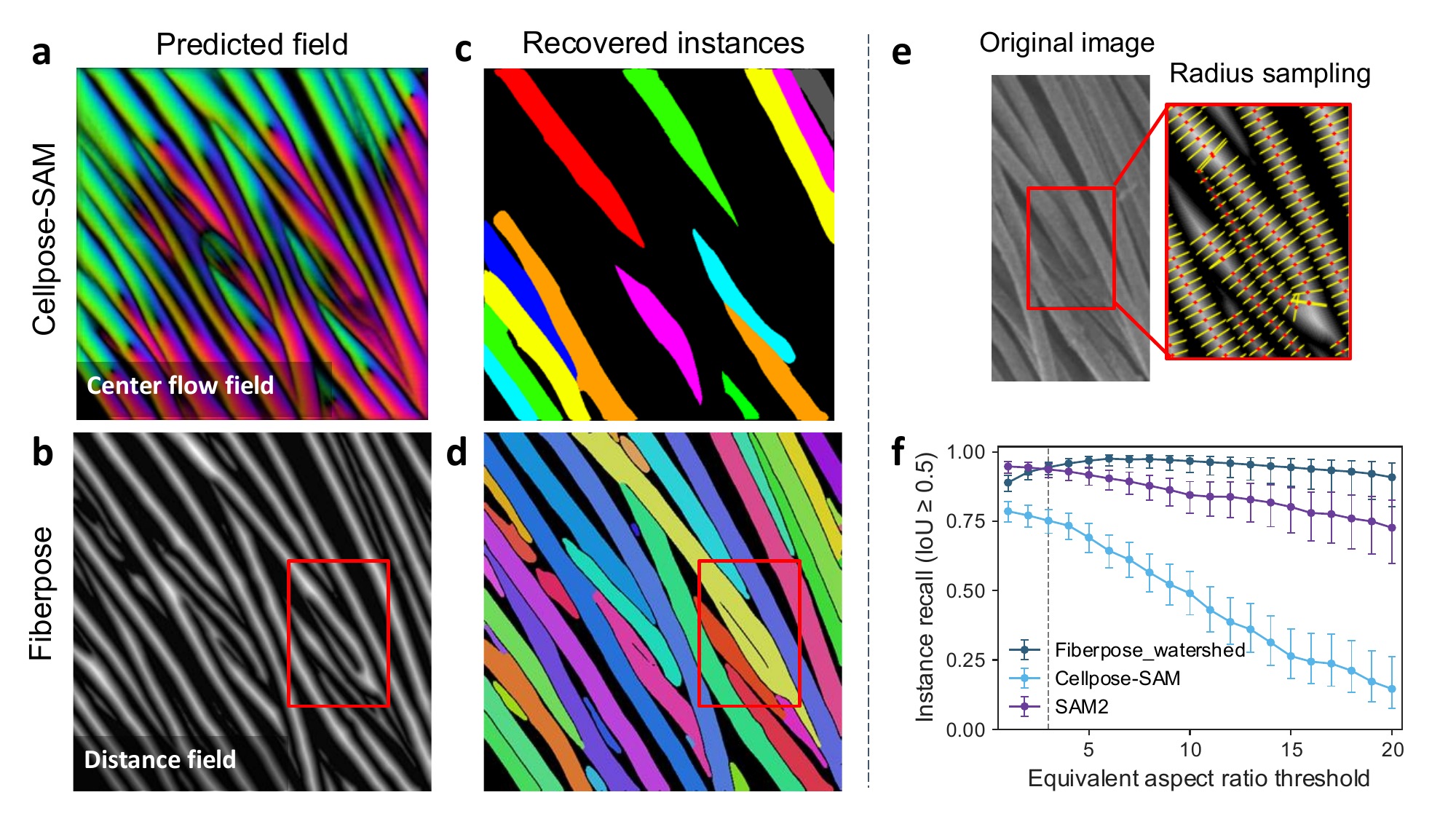}
\caption{\textbf{
Instance recall on elongated objects from different field representations.
}
(a) Cellpose-SAM represents objects using a predicted center-flow field that becomes ambiguous for elongated objects, whereas 
(b) \textit{Fiberpose} predicts a distance field naturally representing long fibers. 
(c,d) Representative instances recovered from the respective fields. Red boxes in (b) and (d) highlight an instance-segmentation error in which several adjacent fibers are merged into a single instance. 
(e) Illustration of local radius sampling within a region affected by this segmentation error. In this example, the predicted distance fields of neighboring fibers come into contact, causing their instance masks to merge. However, the local distance-field ridges remain separable, allowing the fibers to be sampled independently for diameter measurement despite the instance-level error. 
(f) Instance recall (IoU $\geq$ 0.5) as a function of the equivalent aspect-ratio threshold for \textit{Fiberpose} (watershed), Cellpose-SAM, and SAM2, showing that Fiberpose demonstates progressively better recall as more elongated objects are considered.
Source image from \cite{Sedlacek2023SEM}.
}
\label{fig:res_instance}
\end{figure}

\subsection*{Instance segmentation of elongated fibers}

We benchmark \textit{Fiberpose} for instance segmentation on a set of ten SEM images, including five with randomly oriented fibers and five with aligned fibers, and compare it with uncurated SAM2 (automatic masks without filtering; see Methods
)~\cite{kirillov2023sam,ravi2024sam2} and Cellpose-SAM~\cite{Pachitariu2025}. We evaluate two variants of \textit{Fiberpose}, using watershed and ridge-expansion post-processing. To focus on elongated structures, the evaluation is restricted to ground-truth instances with circularity $\leq 0.5$, corresponding approximately to an equivalent aspect ratio $\geq 4$ under a rectangular approximation (see Methods).

Segmentation performance is evaluated using instance recall and precision, together with matched-pair intersection over union (IoU), normalized surface Dice (NSD), and the 95th-percentile Hausdorff distance (HD95) \cite{MaierHein2024}. The image-level metric values are subsequently averaged over the test set. Table~\ref{tab:segmentation_performance} reports the mean and sample standard deviation across the ten test images.

\begin{table}[htbp]
\centering
\caption{\textbf{Comparison of instance segmentation performance.}
Values are reported as mean $\pm$ standard deviation. Only ground-truth instances with circularity $\leq 0.5$ (equivalent aspect ratio $\geq 4$) were evaluated; recall and precision are defined in Methods. Best mean values are highlighted in bold.}
\label{tab:segmentation_performance}
\small
\begin{tabular}{lccccc}
\toprule
\textbf{Method} & \textbf{Recall} & \textbf{Precision} &
\textbf{IoU} & \textbf{NSD} & \textbf{HD95} \\
\midrule
SAM2 (uncurated)
& $0.93 \pm 0.13$
& $0.10 \pm 0.05$
& $0.90 \pm 0.06$
& $0.78 \pm 0.17$
& $32.40 \pm 31.52$ \\

Cellpose-SAM
& $0.70 \pm 0.24$
& $0.90 \pm 0.15$
& $0.91 \pm 0.05$
& $\mathbf{0.79} \pm 0.25$
& $\mathbf{19.40} \pm 16.19$ \\

\hline

Fiberpose (ridge expansion)
& $\mathbf{0.97} \pm 0.06$
& $\mathbf{0.96} \pm 0.03$
& $0.87 \pm 0.05$
& $0.66 \pm 0.20$
& $32.67 \pm 30.81$ \\

Fiberpose (watershed)
& $\mathbf{0.97} \pm 0.05$
& $\mathbf{0.96} \pm 0.02$
& $\mathbf{0.92} \pm 0.02$
& $0.78 \pm 0.15$
& $29.16 \pm 31.33$ \\
\bottomrule
\end{tabular}
\end{table}

\textit{Fiberpose} detected elongated fibers both more completely than Cellpose-SAM and more specifically than SAM2 (Table~\ref{tab:segmentation_performance}). The two baselines failed in complementary ways. Cellpose-SAM rarely produced false instances but missed a substantial fraction of elongated fibers, so its precision remained high while its recall dropped to $0.70$ (for equivalent aspect ratio $\geq 4$). Uncurated SAM2 demonstrated the opposite behaviour: it recovered most fibers, but only alongside a large number of nonspecific object predictions, which reduced its precision to $0.10$. \textit{Fiberpose} combined the strengths of both, reaching the highest recall and the highest precision among the evaluated methods in either post-processing mode. Its recall was significantly higher than that of Cellpose-SAM and its precision significantly higher than that of SAM2 (paired Wilcoxon signed-rank test, $n = 10$ images, $p = 0.002$ for both comparisons and both post-processing modes), while it did not differ significantly from each baseline on that baseline's stronger metric.

This gain in detection did not come at the cost of mask quality. For instances matched to ground truth, IoU, NSD and HD95 of watershed \textit{Fiberpose} did not differ significantly from those of Cellpose-SAM or SAM2, and no method was best on all three metrics (Table~\ref{tab:segmentation_performance}). HD95 also varied substantially across images, which limits interpretability of small differences in its mean. These matched-pair scores must be read together with the detection metrics, because they describe only successfully matched instances and ignore missed or unmatched predictions \cite{Reinke2024}. This distinction is particularly important for uncurated SAM2, which achieves high matched-pair IoU despite very low precision, reflecting a large number of nonspecific predictions.

The choice of post-processing within \textit{Fiberpose} affected boundaries rather than detection. Recall and precision did not differ significantly between the two variants, whereas watershed yielded higher IoU and NSD and lower HD95 than ridge expansion (Table~\ref{tab:segmentation_performance}). Thus, for elongated fibers, \textit{Fiberpose} detected instances more completely than Cellpose-SAM and more specifically than SAM2, with comparable mask quality.

The differences in segmentation performance may be partly explained by the underlying field representations (Fig.~\ref{fig:res_instance}). Cellpose-SAM predicts a center-directed flow field (Fig.~\ref{fig:res_instance}a), which becomes increasingly ambiguous for elongated structures with extended central regions, complicating flow integration and instance reconstruction (Fig.~\ref{fig:res_instance}c). In contrast, \textit{Fiberpose} predicts a distance field (Fig.~\ref{fig:res_instance}b), encoding local distances to fiber boundaries. This representation enables the detection of elongated centerlines through ridge detection and subsequent reconstruction of individual fiber instances (Fig.~\ref{fig:res_instance}d). The difference is further illustrated by instance recall as a function of equivalent aspect ratio (Fig.~\ref{fig:res_instance}f). While Cellpose-SAM recall decreases substantially as progressively more elongated instances are considered, \textit{Fiberpose} maintains consistently high recall, with SAM2 showing intermediate performance.

In rare cases, \textit{Fiberpose} remains susceptible to segmentation errors arising from inaccuracies in the predicted distance field. In particular, artificial connections between adjacent fibers may cause several fibers to merge into a single instance (Fig.~\ref{fig:res_instance}d). However, that instance segmentation and fiber diameter estimation are performed through separate operations on the predicted distance field. Diameters are estimated by sampling local radii along detected centerlines rather than from reconstructed instance boundaries (Fig.~\ref{fig:res_instance}e). Consequently, diameters remain measurable even where adjacent fibers merge into a single instance.

\begin{figure}[htb]
\centering
\includegraphics[width=0.95\linewidth]{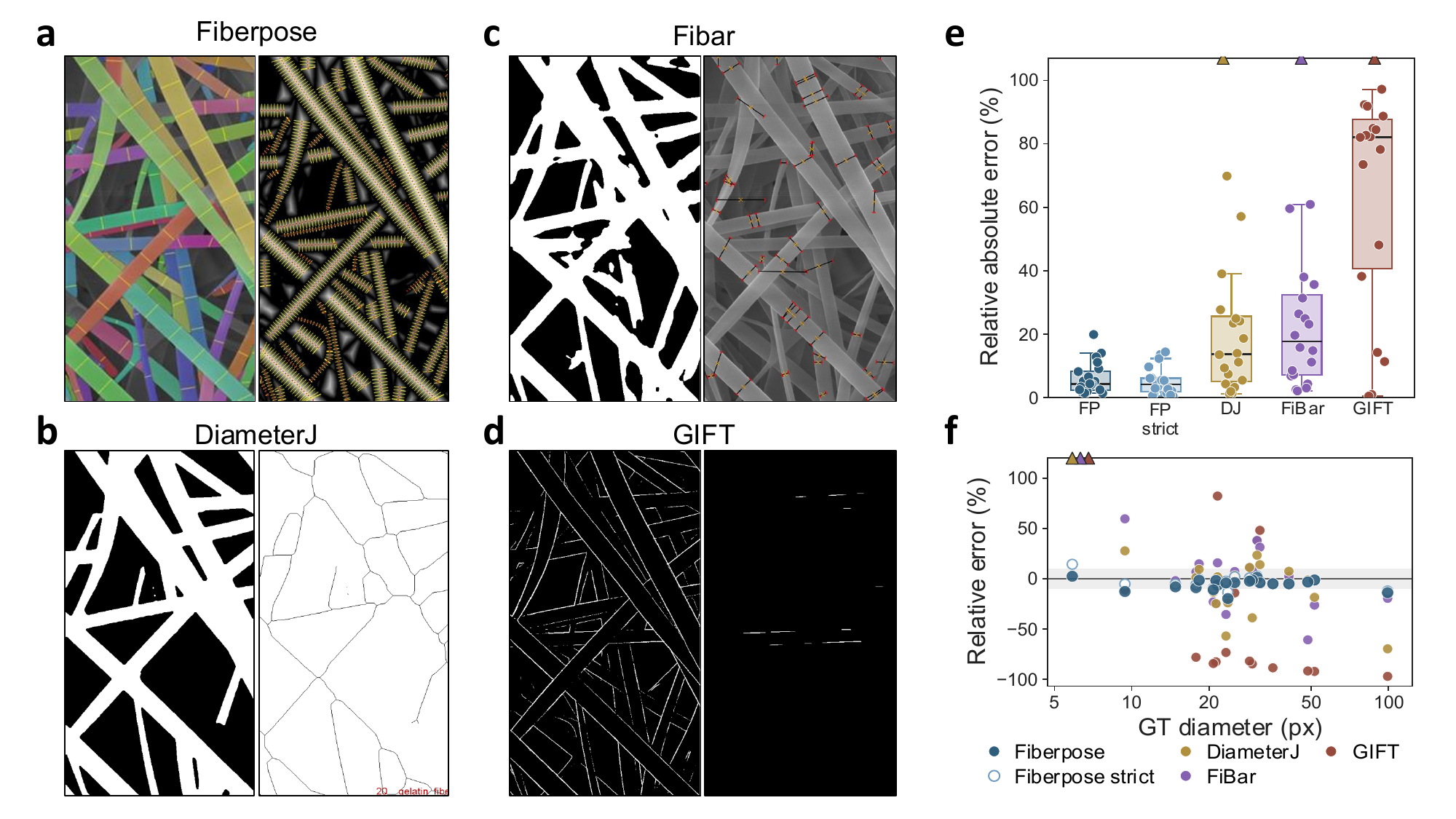}
\caption{\textbf{Comparison of fiber diameter measurements with existing methods.}
(\textbf{a--d}) Outputs of the four methods on a representative SEM image \cite{Huling2022}.
(\textbf{a}) \textit{Fiberpose}: diameter sampling (every tenth line) overlaid on the instance segmentation (left) and full sampling overlaid on the DT field (right).
(\textbf{b}) \textit{DiameterJ}: binary segmentation (left) and the resulting skeleton (right).
(\textbf{c}) \textit{Fibar}: binary segmentation reimplemented following the published pipeline \cite{Moor2025} (left) and measurement marks overlaid on the original SEM image, the only output the tool provides (right).
(\textbf{d}) \textit{GIFT}: edge-detection result (left) and the lines retained after filtering at $0^\circ$ (right).
(\textbf{e}) Relative absolute errors (\%) of the automated diameter estimates with respect to manual measurements. Individual points represent images ($n = 20$; $n = 18$ for \textit{GIFT}, which returned no valid estimate for two images); boxes indicate the median and interquartile range (IQR), whiskers 1.5~IQR. Triangles on the top edge mark values beyond the axis range.
(\textbf{f}) Relative error (\%) of the automated estimates as a function of the manually measured (GT) diameter. The grey band marks $\pm10\%$. Triangles mark values outside the axis range. 
}
\label{fig:res_diameters}
\end{figure}

\subsection*{Diameter accuracy against manual reference}

The diameter benchmark included 20 SEM images not used for training, covering both high-contrast images with clearly defined fibers and challenging images with faint borders, non-uniform fiber contrast and variable sizes. Diameters obtained with \textit{Fiberpose} in default and strict modes, \textit{DiameterJ}~\cite{Hotaling2015}, \textit{Fibar}~\cite{Moor2025} and \textit{GIFT}~\cite{Huling2022} were compared with manual measurements, using the relative absolute error per image (\%) as the metric and the paired Wilcoxon signed-rank test for pairwise comparisons ($n = 20$ images; $n = 18$ for \textit{GIFT}).

\textit{Fiberpose} measured fiber diameters significantly more accurately than all three existing tools. The median relative absolute error was 4.2\% in default mode and 4.1\% in strict mode, compared with 13.7\% for \textit{DiameterJ}, 17.7\% for \textit{Fibar} and 82.2\% for \textit{GIFT} (Fig.~\ref{fig:res_diameters}e), and every tool differed significantly from both \textit{Fiberpose} modes ($p < 0.01$ for all comparisons). \textit{Fiberpose} errors also varied least between images and never exceeded 20\%, over the whole range of fiber diameters (Fig.~\ref{fig:res_diameters}f). \textit{Fiberpose} showed a small systematic underestimation, with a median signed error of $-1.20$~px in default mode and $-0.70$~px in strict mode.

The lower accuracy of \textit{DiameterJ} and \textit{Fibar} did not result from a systematic error. Their median signed errors were close to zero ($-0.20$~px and $+1.10$~px, respectively); instead, both tools were accurate on some images and produced large errors on others, which is reflected in their wide error spread (Fig.~\ref{fig:res_diameters}e,f). These errors can be explained by the quality of the semantic segmentation on which both tools rely: merging of adjacent fibers increased the estimated diameter, whereas incomplete segmentation of a fiber decreased it. The SRM segmentation of \textit{DiameterJ} is susceptible to both errors, and the skeleton derived from it is rough: although \textit{DiameterJ} attempts to filter out fiber junctions, its skeleton remains less precise than the measurement centerlines of \textit{Fiberpose} (Fig.~\ref{fig:res_diameters}a,b). \textit{Fibar} does not provide intermediate segmentation results; a mask reimplemented following its published pipeline contained more incompletely segmented fibers than the \textit{DiameterJ} mask, and the only output of the tool contained many misaligned measurement marks (Fig.~\ref{fig:res_diameters}c). Consequently, both tools provide reliable results only on carefully selected images.

\textit{GIFT} was the most fragile method: it performed acceptably on some images but failed on most of them, and returned no valid estimate for two of the 20 images. Its edge-detection result and the lines retained after filtering at $0^\circ$ show that the underlying signal can be weak and unstable (Fig.~\ref{fig:res_diameters}d). In contrast, \textit{Fiberpose} measures diameters directly on the predicted DT field, and its performance was robust to image appearance (Fig.~\ref{fig:res_diameters}a). Thus, \textit{Fiberpose} was the only method whose diameter error remained low and consistent across the whole benchmark.

\begin{figure}[htb]
\centering
\includegraphics[width=\linewidth]{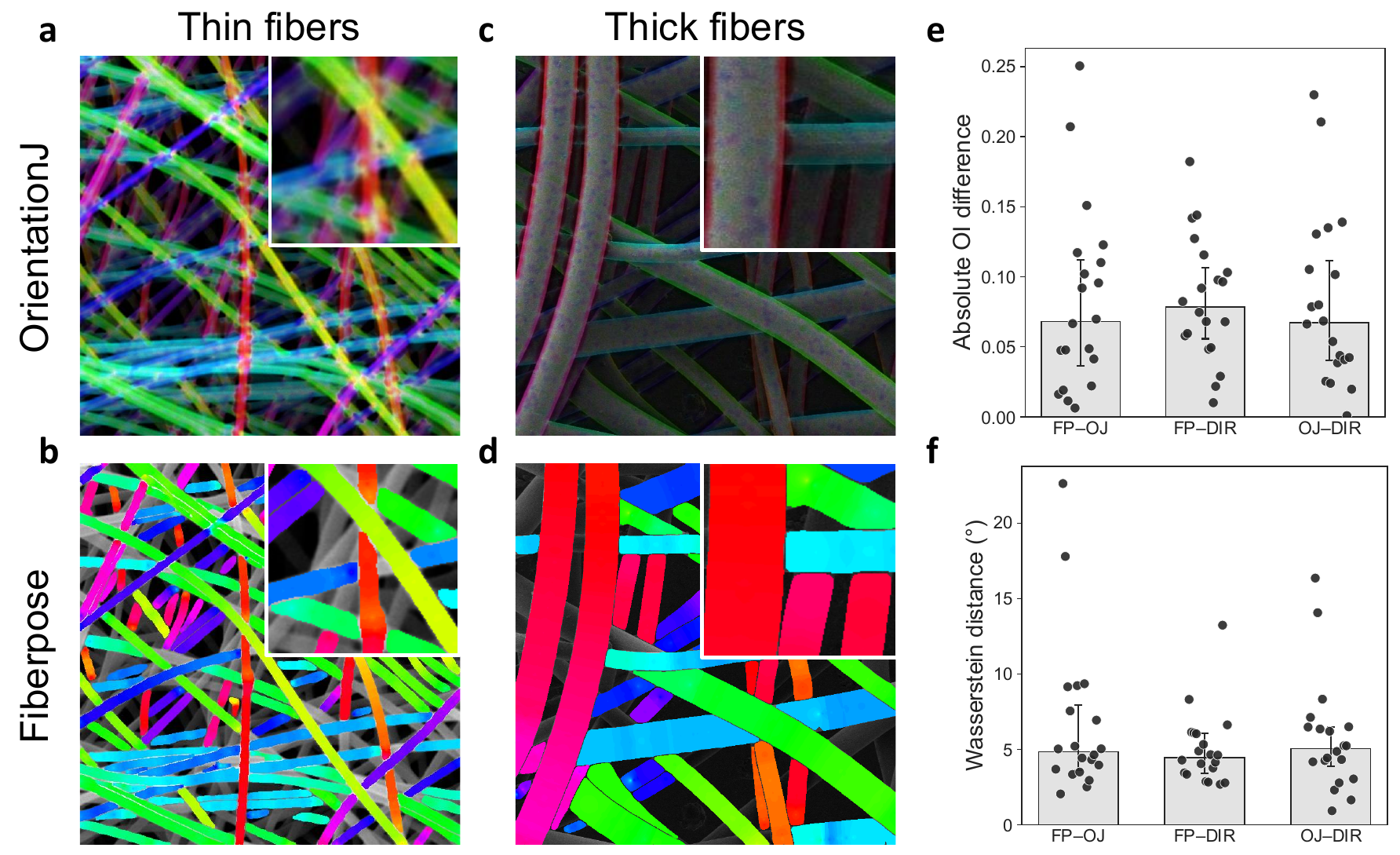}
\caption{\textbf{Comparison of fiber orientation analysis methods.}
(\textbf{a,c}) Representative local orientation maps generated by \textit{OrientationJ} for thin (a) and thick (c) fiber networks.
(\textbf{b,d}) Corresponding orientation maps generated by \textit{Fiberpose}, with local orientations interpolated across segmented fiber regions.
Colors encode local fiber orientations. Insets show magnified views of the corresponding regions, highlighting differences between the methods. In particular, \textit{OrientationJ} primarily captures fiber edges, leaving thick fiber interiors incompletely represented, whereas \textit{Fiberpose} provides orientation estimates across the full fiber width.
Source images from \cite{Uhljar2022, Hotaling2015Dataset}.
(\textbf{e}) Pairwise absolute differences in orientation index (OI) between \textit{Fiberpose} (FP), \textit{OrientationJ} (OJ), and \textit{Directionality} (DIR).
(\textbf{f}) Pairwise Wasserstein distances (in degrees) between the angular orientation distributions obtained with the same three methods. In (e,f), individual points represent images ($n = 20$) and bars indicate the median and interquartile range (IQR). No statistically significant differences were detected between method pairs (paired Wilcoxon signed-rank test, $n = 20$ images).
}
\label{fig:res_orientation}
\end{figure}

\subsection*{Orientation agreement with established tools}

As no manual ground truth is available for fiber orientation, \textit{Fiberpose} was evaluated by its agreement with two widely used tools, \textit{OrientationJ}~\cite{Puspoki2016} and \textit{Directionality}~\cite{Schindelin2012}. Qualitatively, \textit{Fiberpose} and \textit{OrientationJ} produced orientation maps with consistent predominant directions (Fig.~\ref{fig:res_orientation}a--d; color wheel in Fig.~\ref{fig:m_orientation}a). However, \textit{OrientationJ} coloring was blurred and for thick fibers concentrated along their borders, where intensity gradients are strongest, leaving fiber interiors empty (Fig.~\ref{fig:res_orientation}c). \textit{Fiberpose} maps were uniformly filled across the full fiber width and sharply bounded by the segmented instances, independently of fiber thickness (Fig.~\ref{fig:res_orientation}b,d). 

Quantitatively, \textit{Fiberpose} agreed with the established tools as closely as they agreed with each other. Across all images ($n = 20$), the absolute difference in orientation index was similar for all three method pairs, with no statistically significant difference (median $|\Delta\mathrm{OI}| = 0.068$, $0.078$ and $0.067$ for Fiberpose--OrientationJ, Fiberpose--Directionality and OrientationJ--Directionality, respectively; paired Wilcoxon signed-rank test, $p \geq 0.87$).
Circular Wasserstein distances between the angular distributions were also likewise comparable (median $4.8^\circ$, $4.5^\circ$ and $5.1^\circ$; $p \geq 0.13$; Fig.~\ref{fig:res_orientation}e,f). Overall, \textit{Fiberpose} reproduced the orientation statistics of established tools while additionally providing instance-resolved maps that cover the full fiber width.

\begin{figure}[!htbp]
  \centering
  \includegraphics[width=0.9\linewidth]{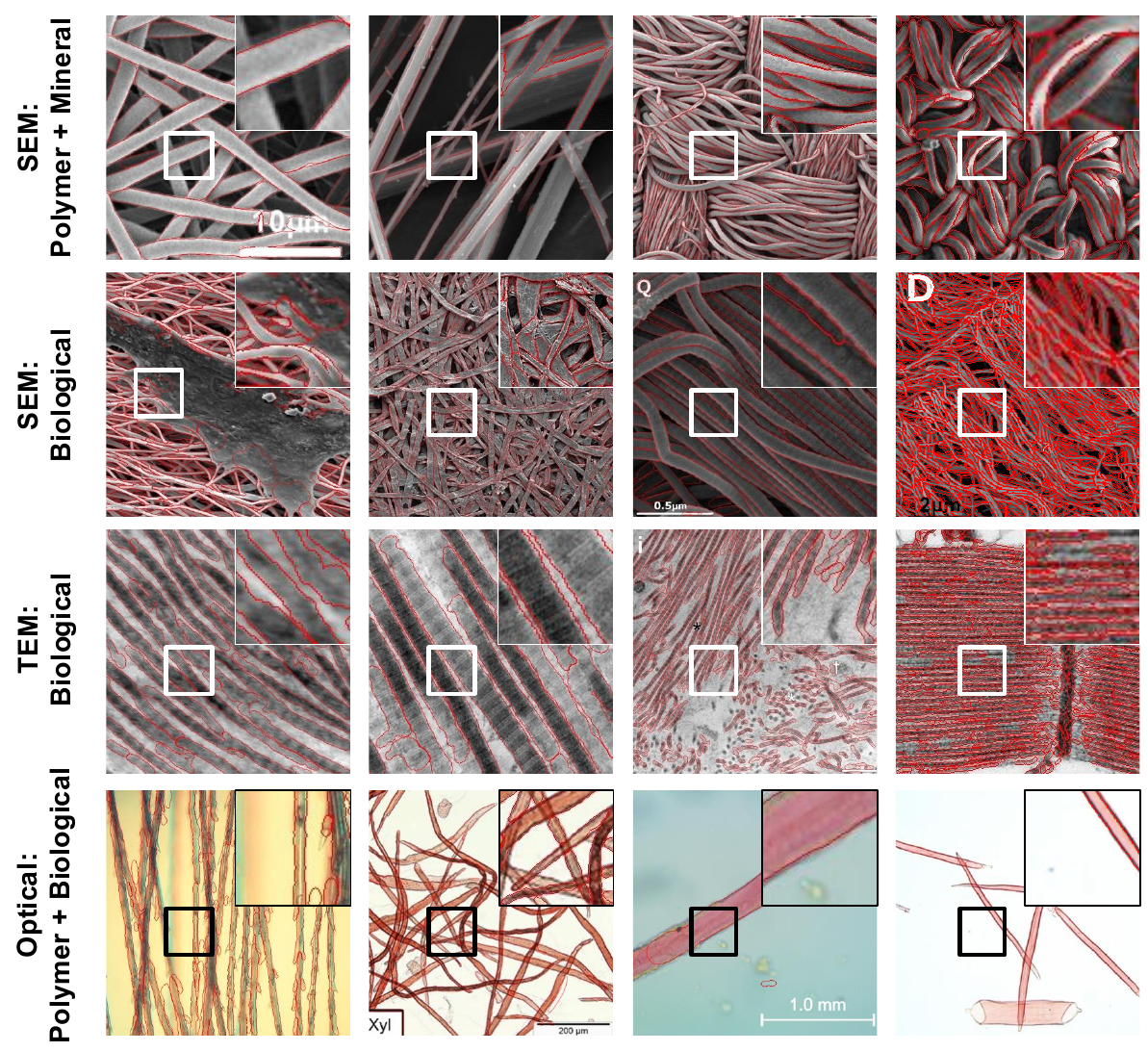}
  \caption{\textbf{Qualitative generalization of \textit{Fiberpose} beyond the training domain.} Each panel shows a source image with \textit{Fiberpose} detections overlaid in red; insets magnify the boxed regions. Samples are listed from left to right.
  \textbf{SEM, polymer and mineral fibers:} electrospun polycaprolactone (PCL) scaffold~\cite{Reid2020}; amosite asbestos fibers~\cite{Ravaka_Amosite}; woven polycotton fabric with polyester and cotton yarns~\cite{Leenders2025}; bundled synthetic filaments of a swimwear fabric~\cite{Atherton2014}.
  \textbf{SEM, biological samples:} cell spread over electrospun PCL+gelatin fibers~\cite{Fee2016}; wood-pulp cellulose fibers in paper~\cite{Kazilek_WoodPulp}; porcine collagen fibrils with periodic banding~\cite{Maurer2018}; collagen-fibril network of a porcine collagen membrane~\cite{Maurer2018}.
  \textbf{TEM, biological samples:} collagen fibrils in native uterine tissue~\cite{Santoso2014}; closely packed collagen fibrils with transverse banding~\cite{Antipova2012}; type I/III collagen fibrils in a human skin equivalent~\cite{Obermayer2026}; myofilaments near a Z-disc in \textit{Drosophila} flight muscle~\cite{GonzalezMorales2017}.
  \textbf{Optical microscopy, plant and animal fibers:} dandelion pappus fibers~\cite{Bachev_Dandelion}; eucalyptus kraft-pulp fibers after xylanase treatment~\cite{Sponla2023}; archaeological animal-hair textile fiber~\cite{CarcamoVega2025}; aspen wood fiber cells beside a vessel element~\cite{Qamar2024}.
  TEM and transmitted-light images were intensity-inverted before inference. Scale bars and letter labels are those of the source figures. All source images were cropped and overlaid with \textit{Fiberpose} detections. They are reproduced under CC BY 3.0/4.0 licences, except the amosite asbestos image (CC BY-SA 3.0) and the wood-pulp paper image (CC BY-SA 4.0); these two panels, as adapted here, are available under CC BY-SA 4.0.}
  \label{fig:generalization}
\end{figure}

\subsection*{Qualitative generalization beyond the training domain}

\textit{Fiberpose} detected fibrous structures in every image group tested, including materials and microscopy modalities absent from its training data (Fig.~\ref{fig:generalization}). The model was trained exclusively on real and synthetic SEM images and was designed to detect bright fibers against a dark background. It was applied here without retraining and with default parameters; the only adaptation was intensity inversion of images acquired using transmission-based modalities, including TEM and conventional transmitted-light microscopy. The examples were taken from openly licensed publications and Wikimedia Commons, and none of the SEM examples belong to the training collection.

The examples are ordered by increasing distance from the training domain. The SEM examples cover polymer and mineral fibers (an electrospun PCL scaffold, amosite asbestos, a woven polycotton fabric and a swimwear fabric) and biological samples (a cell spread over electrospun fibers, wood-pulp paper and porcine collagen fibrils). The TEM examples show collagen fibrils in uterine tissue and in a human skin equivalent, isolated banded collagen fibrils, and myofilaments of \textit{Drosophila} flight muscle. The optical microscopy examples show dandelion pappus fibers, eucalyptus kraft-pulp fibers, an archaeological animal-hair fiber and aspen wood fiber cells. Detections were obtained across all groups despite substantial differences in image contrast, background texture, fiber appearance and scale, from collagen fibrils and myofilaments to textile yarns and wood fibers. Visual inspection indicated a higher frequency of false-positive or incomplete detections in the TEM and optical images, which are not represented in the training set; typical errors were spurious detections in textured background and fragmented outlines of weakly contrasted fibers.

Thus, although trained only on real and synthetic SEM images, \textit{Fiberpose} produced qualitatively meaningful detections across polymer, mineral, plant and animal fibers imaged by electron and optical microscopy. As a next step, we will extend the training dataset to these imaging domains to further improve generalization.

\section*{Discussion}

\textit{Fiberpose} shows that fiber morphometry does not have to pass through a binary mask. By predicting the DT field directly from the image and measuring on that field, it reached a median diameter error of about 4\%, compared with 14--82\% for \textit{DiameterJ}, \textit{Fibar} and \textit{GIFT}, and detected elongated fibers more completely than Cellpose-SAM and more specifically than SAM2. We attribute this to combining the strengths of two lines of work in a single predicted field: the learned geometric field prediction of Cellpose and Omnipose \cite{Cellpose2021, Omnipose2022} and the DT-based measurement of \textit{DiameterJ} \cite{Hotaling2015}.

Three developments were needed to make this approach work. The first is an annotated dataset of fiber SEM images, complemented by synthetic images, which we created for this work. The second is a training procedure that yields a geometrically correct DT field, combining gradient supervision with static and dynamic Eikonal regularization. The third is a pipeline that extracts quantitative data from such predictions: geometric selection of measurement centerlines excludes fiber ends and crossings, where the field is distorted, and two-offset sampling corrects for the rounding of the predicted ridges.

The main consequence of working on the predicted field is that measurement no longer depends on instance segmentation. In our benchmark, the errors of \textit{DiameterJ} and \textit{Fibar} could be traced to their masks, in which merged fibers increased the estimated diameter and incompletely segmented fibers decreased it. \textit{Fiberpose} diameters, read along the measurement centerlines of the DT field, remained measurable even where adjacent fibers merged into a single instance (Fig.~\ref{fig:res_instance}e).

The accuracy of \textit{Fiberpose} diameters stayed consistent across images on which the existing tools ranged from accurate to failing. Orientation results agreed with \textit{OrientationJ} and \textit{Directionality} as closely as these two established tools agreed with each other, while \textit{Fiberpose} additionally provided orientation maps within segmented instances. In terms of mask quality, \textit{Fiberpose} reached matched-instance IoU comparable to Cellpose-SAM, with no statistically significant difference between the two methods. Thus, \textit{Fiberpose} combines mask quality on par with a foundation-model-based segmenter with reliable fiber detection and accurate morphometry.

This was achieved with a compact network. The
pretrained \textit{FPUNet} contains 0.93M parameters (3.7~MB in FP32),
about 7-fold fewer than the models used in the original Cellpose and
Omnipose frameworks ($\sim$6.6M) and about 330-fold fewer than the
transformer-based Cellpose-SAM ($\sim$305M).
Even with the quality-control branch, the complete \textit{FPWNet} remains
compact (1.35M parameters, 5.4~MB). One possible explanation is the relatively constrained geometry of fibrous structures. Unlike cells, which exhibit considerable diversity in shape and internal organization, fibers share characteristic geometric properties, including elongation, locally parallel boundaries, and gradual variations in radius along their length. These properties are largely independent of material composition and imaging modality, potentially reducing the complexity of the features required to infer geometric distance from image intensity. Consequently, a compact convolutional network may be sufficient to learn this mapping without requiring a large number of feature channels. Larger architectures, particularly transformer-based models \cite{Marks2025, Ma2024}, may offer advantages in representational capacity and cross-domain generalization, but these potential benefits must be balanced against increased computational requirements. Systematic benchmarking on larger and more diverse datasets, including measurements of inference time, memory consumption, and accuracy, is needed to determine whether additional model complexity provides a meaningful advantage for DT prediction.

Although the present framework was developed primarily for SEM images, qualitative experiments demonstrated its ability to detect fibrous structures in other imaging domains, including transmission electron and optical microscopy of biological materials. These observations suggest that the model has learned geometric invariants of fibrous structures that generalize across differences in image appearance. However, the model was trained exclusively on real and synthetic SEM images, and reduced detection specificity was observed in previously unseen imaging domains. Broader quantitative validation and training on multimodal datasets will therefore be necessary to establish reliable performance on natural fibers, collagen networks \cite{Bredfeldt2014, Wershof2021}, cytoskeletal structures \cite{Boudaoud2014}, and other biological specimens.

Several limitations remain. Very thin fibers, particularly those only a few pixels wide, are challenging to represent accurately with a smooth distance-transform field because of the limited spatial information available for reliable ridge extraction and gradient estimation. Upscaling may improve numerical stability, although interpolation cannot recover geometric information absent from the original image. Another limitation is that \textit{Fiberpose} identifies individual fiber fragments rather than reconstructing complete fiber networks. Recovering network connectivity would require additional methods, potentially incorporating graph-based approaches \cite{Xu2015, Usov2015} to link fragments and resolve intersections. Finally, the present work focuses on two-dimensional images, although the underlying methodology could be extended to volumetric data through the prediction of three-dimensional distance fields and extraction of spatial centerlines.

In summary, predicting the DT field instead of a binary mask turns a single network output into instance segmentation, diameter and orientation measurements, and on the evaluated SEM images it outperformed existing tools in diameter accuracy and fiber detection. \textit{Fiberpose} is provided as a desktop application and a web tool that displays every intermediate image object, from the DT field to measurement centerlines, so that each measurement can be inspected and reproduced \cite{Laine2021}. With further validation across materials, imaging modalities and laboratories, this common geometric representation could replace the collection of task-specific image-processing procedures currently used for fiber morphometry and support standardized quantitative analysis of fibrous structures.

\section*{Code and data availability}
The source code and pretrained models are available at \url{https://github.com/vikba/fiberpose}. The web tool is available at \url{https://fiberpose.org}. Training used publicly available images and synthetically generated data. The datasets used for training and validation, together with the source publications, licences and attribution information for the public images, are available from the corresponding author upon reasonable request.

\section*{Acknowledgments}
We thank Alexander Minorsky for the help in preparing ground truth images for the training of neural network.

\section*{Declaration of generative AI use}

During the development of the software and preparation of this manuscript, the authors used generative AI tools to assist with software development and language editing. All AI-assisted outputs were reviewed and verified by the authors, who take full responsibility for the scientific content, analyses, and conclusions.

\bibliographystyle{unsrt}
\bibliography{references}

\section*{Technical Methods}

\subsection*{Neural network training}

\paragraph*{Neural network architecture.}
We use a \textit{FPUNet} network based on \cite{Zhang2018ResUNet}, which is a compact residual U-Net encoder--decoder with a configurable number of resolution levels. Each level contains a residual block consisting of two $3\times3$ convolutions with group normalization \cite{Wu2018}, ReLU activation, and a shortcut connection. Downsampling is performed using strided convolutions, while the decoder uses bilinear interpolation and skip connections from corresponding encoder levels. A final $1\times1$ convolution produces the output.
The \textit{FPUNet} comprises five resolution levels with feature widths of 16, 24, 48, 96, and 144 channels and maps a grayscale input image $I(x,y)$ to a single-channel distance-transform field $DT(x,y)$ without an output activation function. 

In addition to the main \textit{FPUNet} network described above, we add an additional second \textit{FPUNet} that estimates foreground probability from the predicted distance field. It is used to identify fiber regions suitable for reliable analysis. This network uses a smaller configuration with four resolution levels and feature widths of 12, 24, 48, and 96 channels, with sigmoid activation applied to its output. When combined, the two networks form \textit{FPWNet} \cite{Jha2020DoubleUNet}, a cascaded architecture in which the distance-transform prediction serves as input to the foreground network.

\paragraph*{Training dataset and annotation.}
A collection of 275 distinct SEM images of fibrous structures was assembled from published scientific articles distributed under CC BY licences. Of these, 60 images were selected for sparse instance annotation and used as real-image training data. The training set additionally included 50 synthetically generated SEM-like images to increase the diversity of fiber geometries and image appearances.

Sparse instance annotation was used \cite{Wolny2022}, distinguishing positive, negative, and ignored regions. Selected fiber fragments were annotated as individual instances, while selected background regions were labeled as negative. All remaining regions were left unlabeled and excluded from direct supervision. Coarse initial masks were generated using Segment Anything 2 \cite{ravi2024sam2} and subsequently refined manually in custom Python script and GIMP image editor. 

Annotated fiber regions were used for supervised distance-transform regression, while trusted background pixels were assigned a target value of zero. Unannotated regions were treated as unknown and excluded from direct regression losses.

\paragraph*{Distance-transform supervision and the training loss.}

The ground-truth distance field $\DT$ was obtained by evaluating \eqref{eq:DT-definition} on the annotated ground truth datasets $\{\Omega_k\}$, with $\DT=0$ outside foreground instances. The network was trained to predict this field, and its output is denoted $\DTp$. All regression terms use the Huber penalty \cite{Huber1964}
\begin{equation}
\rho_\kappa(r)=
\begin{cases}
\tfrac{1}{2}r^{2}, & |r|\le\kappa,\\[2pt]
\kappa\left(|r|-\tfrac{1}{2}\kappa\right), & |r|>\kappa,
\end{cases}
\label{eq:huber}
\end{equation}
with transition parameter $\kappa = 1$, and every loss term is averaged over its own binary mask $M$, with normalizer $N_M=\sum_{x,y}M(x,y)$.

Distance-transform regression was evaluated over the annotated foreground pixels $M_{\mathrm{fg}}$. Where trusted background pixels $M_{\mathrm{bg}}$ were available, an additional term with its own normalization was added, in which the target is zero:
\begin{equation}
\mathcal{L}_{\mathrm{DT}}
=
\frac{1}{N_{\mathrm{fg}}}\sum_{x,y} M_{\mathrm{fg}}(x,y)\,
\rho_\kappa\!\left(\DTp(x,y)-\DT(x,y)\right)
+
\frac{1}{N_{\mathrm{bg}}}\sum_{x,y} M_{\mathrm{bg}}(x,y)\,
\rho_\kappa\!\left(\DTp(x,y)\right).
\label{eq:loss-dt}
\end{equation}
The background term is omitted for images without trusted background ($N_{\mathrm{bg}}=0$).

Spatial derivatives were computed with Scharr operators, $\partial_x f = S_x * f$ and $\partial_y f = S_y * f$, where
\begin{equation}
S_x=\frac{1}{32}
\begin{bmatrix}
-3 & 0 & 3\\
-10 & 0 & 10\\
-3 & 0 & 3
\end{bmatrix},
\qquad S_y = S_x^{\top}.
\label{eq:scharr}
\end{equation}
Gradient supervision penalizes the discrepancy between the spatial derivatives of the predicted and ground-truth fields:
\begin{equation}
\mathcal{L}_{\mathrm{grad}}
=
\frac{1}{N_{\mathrm{fg}}}
\sum_{x,y}
M_{\mathrm{fg}}(x,y)
\Big[
\rho_\kappa\!\left(\partial_x\DTp-\partial_x\DT\right)
+
\rho_\kappa\!\left(\partial_y\DTp-\partial_y\DT\right)
\Big](x,y).
\label{eq:loss-grad}
\end{equation}

To encourage geometric consistency, Eikonal regularization was applied to promote the unit-gradient property of a distance field, $\|\nabla\DT\|_2=1$. The deviation of the prediction from this property is measured by the residual
\begin{equation}
e(x,y)=
\left|
\sqrt{\left(\partial_x\DTp\right)^{2}+\left(\partial_y\DTp\right)^{2}}
-1
\right|.
\label{eq:eik-residual}
\end{equation}
Static Eikonal regularization \cite{Gropp2020} was applied within annotated fiber interiors. Dynamic Eikonal regularization was applied to confidently predicted foreground regions lacking annotations:
\begin{equation}
\mathcal{L}_{\mathrm{Eik,s}}
=
\frac{1}{N_{\mathrm{s}}}\sum_{x,y} M_{\mathrm{s}}(x,y)\,e(x,y),
\qquad
\mathcal{L}_{\mathrm{Eik,d}}
=
\frac{1}{N_{\mathrm{d}}}\sum_{x,y} M_{\mathrm{d}}(x,y)\,e(x,y),
\label{eq:loss-eik}
\end{equation}
with masks
\begin{equation}
M_{\mathrm{s}} = M_{\mathrm{fg}} \wedge \neg R,
\qquad
M_{\mathrm{d}} = \mathbb{1}\!\left[\DTp>\tau\right] \wedge \neg\left(M_{\mathrm{fg}}\vee M_{\mathrm{bg}}\right) \wedge \neg\hat{R},
\label{eq:eik-masks}
\end{equation}
where $\tau = 0.5$ is the confidence threshold. $R$ and $\hat{R}$ are ridge masks of $\DT$ and $\DTp$, respectively, which exclude the regions near distance-field ridges where the distance transform is not differentiable. The dynamic mask $M_{\mathrm{d}}$ is recomputed from the current prediction at every iteration and is treated as a constant during backpropagation.

The combined training objective was
\begin{equation}
\mathcal{L} =
\lambda_{\mathrm{DT}}\mathcal{L}_{\mathrm{DT}}
+\lambda_{\mathrm{grad}}\mathcal{L}_{\mathrm{grad}}
+\lambda_{\mathrm{Eik,s}}\mathcal{L}_{\mathrm{Eik,s}}
+\lambda_{\mathrm{Eik,d}}\mathcal{L}_{\mathrm{Eik,d}},
\label{eq:loss-total}
\end{equation}
where the coefficients control the relative contributions of distance-transform regression, gradient supervision, and static and dynamic Eikonal regularization, respectively.

\paragraph*{Training procedure.}
Input images were converted to grayscale and independently normalized to the range $[0,1]$. During each training epoch, a single random $384\times384$ pixel crop was extracted from each image after applying a random rotation. 
Additional on-the-fly photometric augmentations were applied including variations in brightness, contrast, gamma, non-uniform illumination, Gaussian noise, and blur. Geometric transformations were applied consistently to the input images and distance-transform (DT) targets, whereas photometric augmentations were applied only to the input images.

The network was trained using the AdamW optimizer \cite{Loshchilov2019} in two stages. The initial stage consisted of 300 epochs with a learning rate of $10^{-3}$ and a batch size of 16, using only the Huber loss on predicted DT values. The subsequent refinement stage consisted of 1,700 epochs with a reduced learning rate of $10^{-5}$ and a batch size of 4. During this stage, the loss function combined the Huber loss on DT values, the Huber loss on DT gradients, static and dynamic Eikonal regularization terms, with relative weights of 1, 10, 10, and 1, respectively. Model performance was monitored on a separate validation dataset, 
and the checkpoint minimizing the dynamic Eikonal loss was selected. Training was implemented in PyTorch \cite{Paszke2019}.

\subsection*{Postprocessing}
\paragraph*{Candidate ridge detection and fragment construction.}
Candidate ridges were extracted from the distance-transform (DT) field using a Steger-type subpixel ridge detector \cite{Steger1998}. Gaussian derivatives were computed at $\sigma=1.0$ pixel, with lower and upper ridge-response thresholds of $0.15$ and $0.35$, respectively. Adjacent ridge points with compatible positions and orientations were linked into subpixel ridge polylines.

Local radii $r_i=DT(p_i)$ were sampled from the floating-point DT field at nearest-pixel coordinates. Ridge polylines were grouped into fiber fragments when their rasterized radial expansions of radius $0.5r_i$ overlapped. Each resulting fragment represents a spatially connected, indivisible portion of a fiber or fiber network in the image.

\paragraph*{Measurement centerline selection.}
Measurement centerlines were selected from candidate ridges using four criteria that test how closely the local DT field matches that of an ideal fiber: (i) low variation in local radius, (ii) gradual change in local radius along the ridge, (iii) antiparallel DT gradients on opposite sides of the centerline, and (iv) unit-slope decrease of the DT away from the ridge. Criteria (i) and (ii) were applied by default; criteria (iii) and (iv) were added in the optional strict mode.

Criteria (i) and (ii) were evaluated using sliding windows of $25$ pixels with $50\%$ overlap along each candidate ridge. A window was accepted when the standard deviation of nearest-pixel DT samples was below $0.8$ pixels (criterion i) and the radius-change angle satisfied (criterion ii)

\begin{equation}
\varphi =
\arctan\left(
\frac{|DT(p_{\mathrm{end}})-DT(p_{\mathrm{start}})|}
{L}
\right),
\qquad \varphi < 30^\circ,
\end{equation}

where $p_{\mathrm{start}}$ and $p_{\mathrm{end}}$ are the first and last ridge points of the window and $L$ is the arc length between them. Overlapping accepted windows were merged into measurement centerlines, and sections shorter than $25$ pixels were discarded. FiberFragments without accepted measurement centerlines remained eligible for instance segmentation.

In strict mode, DT values and gradients were additionally sampled at each centerline point on both sides of the fiber, at offsets of $\pm r_i/2$ along the centerline normal. Criterion (iii) was met when the negative dot product of the two normalized gradients was at least $0.95$, indicating approximately antiparallel directions. Criterion (iv) was met when both sampled DT values deviated by no more than $0.1r_i$ from $r_i/2$, the value expected at this offset for a DT field of unit slope. A measurement centerline section was retained when at least $70\%$ of its evaluated points met both criteria. Strict mode was disabled by default in the Fiberpose application.

\paragraph*{Instance segmentation.}
Two alternative methods were implemented to reconstruct fiber instances from candidate ridges grouped into fiber fragments (Figure~\ref{fig:segmentation_mech}).

\textit{Marker-controlled watershed \cite{Beucher2018, Chen2004}:}
The candidate ridges belonging to retained fiber fragments were rasterized as one-pixel-wide markers, with a unique label assigned to each fragment. The distance transform was normalized as $DT_n=DT/255$ and converted into a watershed elevation image $E=1-DT_n$. Segmentation was restricted to a mask obtained by thresholding distance-field gradient magnitude, $\|\mathbf{g}\|\geq g_{\min}$ (default 0.5), and adding the centerline mask dilated twice using a $3\times3$ square kernel.  Watershed segmentation was performed within this mask using eight-neighbor connectivity, allowing regions to grow outward from the centerlines while preserving separate fragment labels.

\textit{Ridge expansion:}
Alternatively, each centerline point was expanded into a disk with radius $R_i=\alpha DT(p_i)$, where $\alpha$ is the expansion factor ($1.0$ by default). The union of disks belonging to the same fragment defined its segmented region (Fig.~\ref{fig:segmentation_mech}).

\paragraph*{Quality control and pruning.}
Before instance reconstruction, fragments were retained only if their longest constituent centerline was at least $3$ pixels long. The remaining fragments were ranked by total accepted measurement-centerline length, and the lowest-ranked $1\%$ fragments were removed.

Gradient quality was assessed within the full-radius expansion of each fragment, excluding centerline pixels, using the median deviation of the smoothed distance-field gradient magnitude from unity. Fragments with fewer than eight valid evaluation pixels were discarded, while additional rank-based filtering by gradient quality was disabled.

After instance reconstruction, an optional foreground-probability map was used to remove false-positive segments. Regions with a mean foreground probability below $0.5$ were discarded. This final filtering was applied identically to both segmentation methods and omitted when no foreground-probability output was available.

\subsection*{Quantitative analysis}
\paragraph*{Local diameter and orientation.}
Centerlines were sampled for measurements at a user-defined spacing. At each sampling point $\mathbf p_i$, the initial fiber normal was estimated from the local centerline tangent and refined over three iterations using distance-transform gradients sampled on opposite sides of the centerline, at half the central distance-transform value. Normal directions were adjusted to maintain continuity along each centerline.

The local radius was estimated with Eq.~\eqref{eq:two_offset}, using distance-transform values sampled at $\delta = 1$~px on both sides of the centerline. Local fiber orientation was calculated perpendicular to the refined normal and expressed in the axial interval $[-90^\circ,90^\circ]$. The resulting diameter and orientation measurements were used to calculate distributions and summary statistics.

\paragraph{Comparison of apex and corrected estimates.}
The comparison used the ten SEM images of the instance-segmentation benchmark described below (five with randomly oriented and five with aligned fibers; none used for training). In each image, ten fiber diameters were measured in ImageJ with the straight-line tool, drawing each line from edge to edge across the fiber, perpendicular to its axis, on clearly identifiable fiber segments away from crossings and fiber ends. Line length, angle and bounding rectangle were exported to perform automatic DT measurements at the same locations. Because Fiberpose samples on the DT ridge, the sampling center $\mathbf{c}$ was set to the position of the maximal DT value along the line, and the unit vector $\mathbf{n}$ along the line was used as the sampling direction. The apex estimate was calculated as $d = 2D(\mathbf{c})$ and the corrected estimate with Eq.~\eqref{eq:two_offset} at $\mathbf{x} = \mathbf{c}$ and $\delta = 1$~px. Each sampling position was rounded to the nearest pixel, without interpolation. Errors were computed as DT-derived minus manual diameter and summarized per image as bias and MAE. Images were treated as the experimental unit: per-image MAE of the apex and corrected estimates were compared with a one-sided paired Wilcoxon signed-rank test ($n = 10$ images); the test was one-sided because the correction is designed to remove the underestimation of the apex estimate. The 95\% confidence interval of the residual bias of the corrected estimate was calculated from the ten per-image mean biases, using the $t$ distribution with 9 degrees of freedom. The perpendicularity of the manual lines was checked against the local fiber normal.

For display in Fig.~\ref{fig:res_diameter_sampling}a, the boundary of the predicted DT field was drawn as the $\mathrm{DT} = 0.5$~px contour, after removing connected components smaller than 0.2\% of the image area. Crops were chosen automatically as the square region containing the largest number of complete manual lines.

\paragraph*{Orientation statistics.}
The orientation index (OI) was calculated from the sampled fiber orientations as

\begin{equation}
\mathrm{OI}
=
\sqrt{
\left\langle\cos(2\theta)\right\rangle^2+
\left\langle\sin(2\theta)\right\rangle^2
}.
\end{equation}

The doubled angles account for the axial nature of fiber orientation, where directions separated by $180^\circ$ are equivalent. OI ranges from zero to one, with higher values indicating stronger overall alignment and lower values indicating greater angular dispersion.

\paragraph*{Orientation map visualization.}
Dense orientation maps were generated by interpolating sampled orientations within each segmented fiber fragment. Each segmented pixel was colored according to an orientation interpolated from up to four nearest centerline samples within the same fragment, using inverse-distance weighting and axial-angle averaging.

\begin{equation}
\theta(\mathbf{x})
=
\frac{1}{2}\operatorname{atan2}
\left(
\sum_i w_i\sin(2\theta_i),
\sum_i w_i\cos(2\theta_i)
\right),
\qquad
w_i=\frac{1}{\|\mathbf{x}-\mathbf{p}_i\|+\varepsilon},
\end{equation}
where $\varepsilon=10^{-6}$ prevents division by zero and distances are measured in pixels. Fragments without valid orientation samples were excluded. The resulting orientation maps were visualized using a cyclic HSV color map.

\paragraph{Instance-segmentation accuracy.}
 Instance segmentation was evaluated on ten SEM images (512 $\times$ 512 pixels) of fibers, including five images with randomly oriented fibers and five with aligned fibers. The images were drawn from public sources: the \textit{DiameterJ} \cite{Hotaling2015Dataset} and \textit{GIFT} \cite{Huling2022} datasets, a collection of SEM images of climbing ropes \cite{Sedlacek2023SEM}, and four open-access articles \cite{Roberto2022, Ciszewski2019, Bloise2018, Shaker2023}. None of the benchmark images were used for training. The test images were manually annotated with three classes: fiber instances, background, and ignore regions. \textit{Fiberpose} in watershed and ridge-expansion segmentation modes was compared with Cellpose-SAM \cite{Pachitariu2025}, used as distributed on Hugging Face, and with uncurated outputs of the SAM2 tiny model \cite{kirillov2023sam, ravi2024sam2} in automatic mask generation mode.

Predicted and ground-truth (GT) instances were first matched one-to-one at an intersection-over-union (IoU) threshold of 0.5, maximizing the number of accepted matches and then their total IoU. Matching preceded all size and shape filtering. The filter was applied to GT instances only: GT instances with area $\geq10$ pixels
and circularity $C\leq0.5$ were retained for evaluation, where
\[
C=\frac{4\pi A}{P_{\mathrm{C}}^2},
\]
with $A$ denoting instance area and $P_{\mathrm{C}}$ the Crofton
perimeter estimate \cite{vanderWalt2014}. IoU, normalized surface Dice (2-pixel tolerance),
and the 95th-percentile Hausdorff distance were then calculated for
retained matched pairs and averaged within each image. For instance
recall and precision, true positives (TP) were predictions matched to
retained GT instances, false negatives (FN) were retained GT instances
without a match, and false positives (FP) were predictions without an accepted match to any
GT instance, provided that less than $50\%$ of their pixels lay inside
annotated ignore regions and that their area after removal of ignored
pixels was at least $30$ pixels. No circularity filter was applied to
predictions, and duplicate predictions of the same fiber also counted as FP. Recall was
calculated as $\mathrm{TP}/(\mathrm{TP}+\mathrm{FN})$ and precision as
$\mathrm{TP}/(\mathrm{TP}+\mathrm{FP})$ in each image. Predictions matched to
excluded GT instances were counted neither as TP nor as FP, and annotated
ignore regions were excluded from evaluation.

For shape-dependent recall analysis, circularity was converted to
rectangular-equivalent aspect ratio:
\[
\mathrm{AR}_{\mathrm{eq}}
=\frac{\pi-2C+\sqrt{\pi(\pi-4C)}}{2C},
\qquad 0<C\leq\frac{\pi}{4}.
\]
Objects with $C>\pi/4$ were assigned $\mathrm{AR}_{\mathrm{eq}}=1$.
Using fixed matches and all GT instances with area $\geq10$ pixels,
cumulative recall was calculated across images for
$\mathrm{AR}_{\mathrm{eq}}\geq t$ at successive cutoffs $t$,
with 95\% Wilson intervals \cite{Wilson1927}.

\paragraph{Diameter-measurement accuracy.}
Diameter-measurement accuracy was evaluated on 20 SEM images ranging in size from $384\times302$ to $1584\times1455$~pixels. The benchmark included both high-quality, high-contrast images with clearly defined fibers and more challenging images characterized by faint boundaries, non-uniform contrast, blur, or reduced image quality. The images were drawn from public sources: the \textit{DiameterJ} \cite{Hotaling2015Dataset} and \textit{GIFT} \cite{Huling2022} datasets and open-access articles \cite{Uhljar2022, Saporito2018, MadejKielbik2022, Huang2019, Atherton2014, Watson2022, Chavoshnejad2020}. None of the benchmark images were used for training.

Measurements obtained with \textit{Fiberpose} in its default and strict modes were compared with those obtained using \textit{DiameterJ} \cite{Hotaling2015}, \textit{Fibar} \cite{Moor2025} and \textit{GIFT} \cite{Huling2022}, with manual measurements serving as the reference. Manual diameter measurements were performed in ImageJ using an overlaid grid consisting of three horizontal lines (top, middle, and bottom), each containing ten sampling points. At each point, the diameter was measured on a clearly identifiable segment of the fiber whose boundary was closest to the sampling point, yielding 30 measurements per image. For each image, the relative absolute error (\%) between the diameter estimated by each automated method and the manually measured median diameter was calculated and compared across methods. \textit{Fiberpose}, \textit{DiameterJ} and \textit{Fibar} estimates were per-image medians, whereas \textit{GIFT} reports the mean of a Gaussian distribution fitted to its diameter measurements; this value was used as reported by the tool. \textit{DiameterJ} and \textit{Fibar} were run with their default settings. \textit{GIFT} was run with filtering angles of $0^\circ$, $45^\circ$, $90^\circ$ and $135^\circ$. Negative diameters returned by \textit{GIFT} (two images) were treated as failed fits and excluded from the comparisons.

Because \textit{Fibar} does not export intermediate segmentation results, its segmentation was reimplemented for visualization following the classical image-processing pipeline described by Moor et al.~\cite{Moor2025}. Grayscale SEM images were denoised with a $3\times3$ Kuwahara filter, binarized with Huang's automatic thresholding method and post-processed with a $15\times15$ median filter to obtain binary fiber masks. The reimplemented masks were used only for illustration (Fig.~\ref{fig:res_diameters}c); all \textit{Fibar} diameter estimates were obtained with the original tool.

\paragraph{Orientation agreement.}
Orientation measurements on the same 20-image morphometry benchmark were compared with those produced by \textit{OrientationJ} \cite{Puspoki2016} and \textit{Directionality} \cite{Schindelin2012}. Both tools were run with their default settings. No manual orientation ground truth was generated. All methods used the same axial-angle convention and histogram binning. Agreement was evaluated using differences in orientation index and Wasserstein distances between the corresponding angular distributions.

\paragraph*{Statistical analysis.}
Images were treated as independent experimental units, with measurements paired by image identifier. All comparisons between methods used paired Wilcoxon signed-rank tests on per-image values, implemented in SciPy \cite{Virtanen2020}: instance-segmentation recall, precision, IoU, NSD and HD95 ($n = 10$ images; exact test); relative absolute diameter errors ($n = 20$ images; $n = 18$ for \textit{GIFT}); orientation-agreement measures (absolute OI differences and Wasserstein distances) between method pairs ($n = 20$ images); and per-image MAE of the apex and corrected estimates ($n = 10$ images; exact test). Tests were two-sided, except for the comparison of apex and corrected estimates, which was one-sided as described above. Statistical significance was defined as $p<0.05$, without correction for multiple comparisons.

\subsection*{Software implementation}
The reference implementation is in Python (PyTorch, NumPy, OpenCV, scikit-image) with an optional PyQt6 GUI for interactive use and batch/CLI for headless runs. The GUI supports GPU inference, DT visualization, segmentation overlays, diameter/orientation histograms, and batch processing, see Figure~\ref{fig:GUI}. The source code with installation instructions is freely available on GitHub  \url{https://github.com/vikba/fiberpose}

\end{document}